\documentclass[10pt,twocolumn,letterpaper]{article}

\usepackage{iccv}              
\usepackage{times}
\usepackage{epsfig}
\usepackage{graphicx}
\usepackage{amsmath}
\usepackage{amssymb}
\usepackage{amsmath,amssymb,amsfonts}
\usepackage{amsmath}
\usepackage{amssymb}
\usepackage{float} 
\usepackage{multirow}
\usepackage{algorithm}
\usepackage{algpseudocode}
\usepackage{subcaption}
\usepackage{graphicx}
\usepackage{colortbl}
\usepackage{tabularx}
\usepackage{makecell}
\usepackage{booktabs}

\usepackage{placeins}

\definecolor{iccvblue}{rgb}{0.21,0.49,0.74}
\usepackage[pagebackref,breaklinks,colorlinks,allcolors=iccvblue]{hyperref}

\def\paperID{*****} 
\def\confName{ICCV}
\def\confYear{2025}

\title{MoPEQ: Mixture of Mixed Precision Quantized Experts}

\author{
Krishna Teja Chitty-Venkata\textsuperscript{1},
Jie Ye\textsuperscript{2},
Murali Emani\textsuperscript{1}\\
\textsuperscript{1}Argonne National Laboratory,
\textsuperscript{2}Illinois Institute of Technology\\
{\tt\small schittyvenkata@anl.gov, jye20@hawk.iit.edu, memani@anl.gov}
}

\begin{document}
\maketitle
\begin{abstract}
Large Language and Vision Models using a Mixture-of-Experts (MoE) architecture pose significant challenges for deployment due to their computational and memory demands. Mixed Precision Quantization assigns different precisions to different layers of an LLM/VLM based on layer sensitivity and importance within the model. In this work, we propose a Post Training Quantization algorithm, MoPEQ, that assigns optimal bit width to each expert. Our method balances accuracy and model size by analyzing each expert's sensitivity using Hessian trace approximation instead of relying on the activation frequency of the expert. This per-expert granularity approach clusters similar experts to maintain model performance while reducing memory requirements. The experimental results on VLMEvalKit benchmark datasets using State-of-the-art VLMs Deepseek-VL2 -tiny, -small, -base, and MolmoE models demonstrate that our mixed precision quantized MoEs achieve competitive accuracy with substantial improvements in memory footprint compared to uniform-precision baseline methods. We perform a comprehensive study to analyze the impact of expert activation frequency and sensitivity using Hessian trace approximation at both layer-wise and model-wide expert precision allocation of 2, 3, and 4 bits to provide a thorough understanding of mixed precision quantization of VLM-MoEs. The code is available \href{https://github.com/krishnateja95/MoE-Mixed-Prec}{\textcolor{blue}{here}}.

\end{abstract}

\section{Introduction}


Large Language Models (LLMs), such as GPT \cite{brown2020language}, PaLM \cite{chowdhery2023palm}, Deepseek-V3 \cite{liu2024deepseekv3} and LLaMA \cite{touvron2023llama}, are based on transformer architectures trained on vast text data to learn and generate human-like language. These networks achieve remarkable performance in tasks such as text completion and translation using hundreds of billions of parameters and attention mechanisms. Vision-Language Models (VLMs) \cite{zhang2024vision} integrate visual and text understanding to tackle multimodal tasks such as image-text retrieval, visual question answering (VQA), and image captioning. These models employ cross-modal transformers to align visual features with text representations, enabling tasks requiring reasoning over images/videos and text.

\begin{figure}[H]
    \centering
     \includegraphics[width=\linewidth]{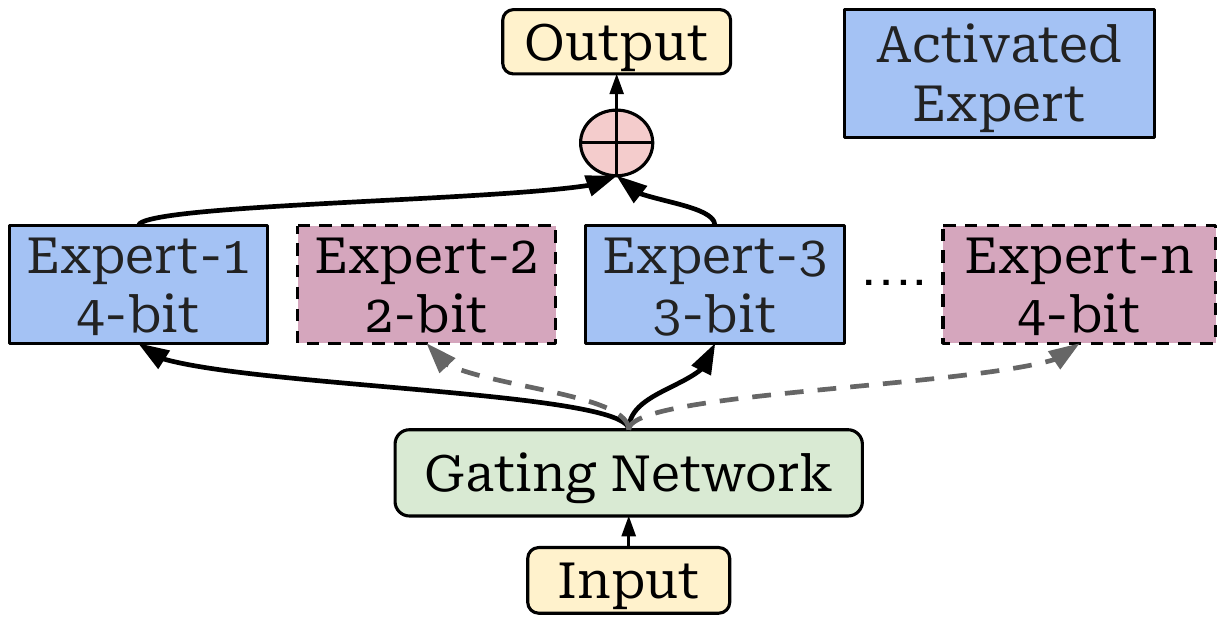}
        \caption{MoPEQ: Mixed Precision Quantization of MoE}
        \vspace{-4mm}
        \label{fig:MoPEQ}
    \captionsetup{justification=centering}
\end{figure}

A Mixture of Expert (MoE) \cite{cai2024survey, liu2024survey, shazeer2017outrageously}, such as Deepseek-MoE \cite{dai2024deepseekmoe}, is a type of network where multiple specialized subnetworks (or experts in general) exist within the same network. The experts are activated through a learned gating network. This kind of architecture allows the execution of complex tasks by dividing them among smaller, specialized subnetworks. Unlike traditional dense models, MoE experts are not pre-defined for specific domains but instead develop their specializations organically during training. The gating network plays a key role by activating only the most suitable experts based on the input tokens. This type of approach enhances computational efficiency and improves task performance by leveraging the strengths of each expert.

LLM Quantization \cite{zhu2024survey} is an efficient method to address the storage challenges imposed by the immense computational and memory demands of LLMs and VLMs. The model size can be decreased by reducing the numerical precision of parameters, accelerating inference while maintaining model accuracy \cite{wan2023efficient}. There are two approaches to quantize a pretrained LLM, which are as follows: (i) Post Training Quantization (PTQ) is a method to reduce precision after training the model, and (ii) Quantization Aware Training (QAT) simulates the low-precision effects during training to mitigate accuracy loss, although requiring an additional cost. PTQ methods (Eg: GPTQ \cite{frantar2022gptq} and AWQ \cite{lin2024awq}) and QAT methods (Eg: QAT-LLM \cite{liu2023llm}) demonstrate robust results, compressing models by up to 4$\times$ with minimal accuracy loss. Mixed precision quantization can optimize LLMs by assigning different precisions to different layers to preserve accuracy. This approach can balance hardware efficiency and accuracy by assigning higher precision to critical parts of the model while using lower precision for less sensitive components. Mixed precision quantization is particularly well-suited for MoE models due to their unique architecture, where only a subset of experts is activated for any given input. This selective activation enables targeted compression strategies to reduce the size of MoE models as uniform quantization can lead to suboptimal compression.



\textbf{Challenges and limitations of State-of-the-art Methods:} The problem of MoE mixed precision quantization for Vision Language models remains underexplored, although there are a few works in the LLM space.
A common approach in the literature to quantize experts in MoE models is to allocate precision based on the expert activation frequency \cite{huang2024mc, li2024examining}. 
Specifically, experts that are activated more frequently are assigned higher bit widths. However, this method does not account for each expert's sensitivity to quantization. An expert with low activation frequency is not necessarily inactive throughout the inference. In scenarios where such experts are used for generation, quantization-induced errors can propagate throughout the network, especially if the expert is highly sensitive to quantization. 
Additionally, models like DeepSeek-V2 \cite{wu2024deepseek} incorporate load balancing loss during training to ensure uniform utilization of all experts, making it challenging to differentiate between the most and least important experts.
Computing activation frequency also requires a calibration dataset, which may not always be representative of all the diverse and varying tasks.
Calibrating with limited data samples to capture activation frequency can tune the quantization to a specific dataset but may limit its generalizability across diverse tasks.

\textbf{MoPEQ:} To address the challenges of MoE quantization, we propose MoPEQ, an efficient algorithm for allocating different bit widths to experts (illustrated in Figure \ref{fig:MoPEQ}). Our approach leverages Hessian trace approximation to estimate expert sensitivity without relying on a dataset. Unlike previous layer-wise assignment methods that focus only on mixed precision within a layer, we allocate higher bit width to layer with high sensitivity.
We construct a mixed-precision search space and employ an expert clustering method to group similar experts based on their relative importance.\looseness=-1



\textbf{Contributions} The main contributions of our paper are: 

    \textbf{(1)} We introduce \textbf{MoPEQ}, a mixed-precision assignment algorithm for allocating varying bit widths to different experts in MoE layers. To the best of our knowledge, there are no published works on mixed-precision quantization for VLM-MoEs. We investigate the impact of mixed precision expert quantization on vision and language token scenarios.  
    
    \textbf{(2)} We compare our expert mixed precision assignment method based on expert sensitivity with the activation frequency approach \cite{li2024examining} and layer-wise expert precision assignment \cite{huang2024mc}. We show that the sensitivity of each expert plays a bigger role than activation frequency. We also demonstrate that our sensitivity-based approach can distinguish the importance of experts and reduce the model size where activation frequency is uniform across layers. In this work, we limit our mixed precision scope only to experts in MoE layers and other layers are quantized uniformly.
    
    \textbf{(3)} We evaluate our quantization approach on four state-of-the-art VLMs: \textit{MolmoE-1B} \cite{deitke2024molmo} and \textit{DeepSeek-VL2} (tiny, small, and base variants) \cite{wu2024deepseek}, using multiple VLMEvalKit tasks \cite{duan2024vlmevalkit}. Our results demonstrate a model size reduction of approximately $1.5\times$ while maintaining accuracy within 5\% across seven tasks.

\section{Background}

\subsection{Mixture of Expert (MoE)}
A Mixture of Expert architecture redesigns the traditional dense network by decomposing monolithic layers into specialized subnetworks called \textit{experts} $E_1, E_2, \ldots, E_N$. These experts are chosen per token based on a trainable gating mechanism $G(x)$. For an input token embedding $\mathbf{x} \in \mathbb{R}^d$, the gate computes normalized weights $\mathbf{g} = \text{Softmax}(W_g\mathbf{x} + \epsilon)$ where $W_g \in \mathbb{R}^{N \times d}$ is the routing matrix and $\epsilon \sim \mathcal{N}(0, \sigma^2)$ induces exploration noise for load balancing. SOTA MoE models, such as Mixtral 8x7b \cite{jiang2024mixtral} and DeepSeek-R1 \cite{guo2025deepseek}, employ \textbf{sparse top-$k$ routing}, choosing only $k$ experts per token through $\text{TopK}(\mathbf{g}, k)$, reducing the computational complexity of the layer to $O(kd^2)$ from $O(Nd^2)$ for dense layers. The final output combines expert contributions through $\mathbf{y} = \sum_{i \in \text{TopK}(\mathbf{g},k)} g_i \cdot \text{FFN}i(\mathbf{x})$, while using an auxiliary loss $\mathcal{L}{\text{aux}} = \lambda \cdot \text{CV}(\text{Load})$ to enforce balanced expert utilization by penalizing the coefficient of variation in token assignments. This kind of expert-based architecture enables parameter scaling to a trillion parameters while minimizing the inference costs using dynamic sparsity. However, there are still challenges associated with balancing experts. Recent advances in MoE-based architecture for the Vision Language Model include DeepSeek-VL2 \cite{wu2024deepseek}, MolmoE-1B \cite{deitke2024molmo}, and LLaVA-MoE \cite{lin2024moe}.

\subsection{Quantization}

Quantization reduces the computational cost of LLMs by approximating high-precision tensors with lower precision. Traditional QAT methods require fine-tuning network weights using the full training set, which may not be feasible in a few cases due to the lack of access to the datasets. PTQ utilizes small calibration sets to optimize rounding thresholds and minimize the $\ell_2$-distance between quantized and full-precision matrices. 
For LLMs, dynamic search techniques optimize rounding functions across varying activation distributions, addressing outliers through transformations like SmoothQuant \cite{xiao2023smoothquant} and ZeroQuant \cite{yao2022zeroquant}. VLMs introduce additional complexity as they integrate visual encoders (e.g., ViT, CLIP) with language modules. Mixed-precision quantization optimizes the trade-off between model compression and accuracy by dynamically assigning bit widths to the neural network components based on their contribution to the task performance.

\subsection{Quantization Methods} \label{sec:quant_methods}

\textbf{GPTQ} \cite{frantar2022gptq} is a layer-wise quantization method that minimizes the reconstruction error between the original and quantized outputs by using second-order information. The objective is to minimize $\min_{\hat{\mathbf{W}}} \left| \mathbf{W}\mathbf{X}^\top - \hat{\mathbf{W}}\mathbf{X}^\top \right|_F^2$, reformulated using the Hessian matrix $\mathbf{H}$ to minimize $\text{Trace}\left( (\mathbf{W} - \hat{\mathbf{W}})^\top \mathbf{H} (\mathbf{W} - \hat{\mathbf{W}}) \right)$. GPTQ accelerates this by using Cholesky decomposition, reducing complexity from $\mathcal{O}(n^3)$ to $\mathcal{O}(n^2)$ per layer, and quantizing weights in groups to reduce memory overhead. \textbf{AWQ} \cite{lin2024awq} introduces an activation-aware quantization method that preserves model accuracy by identifying and protecting salient weights using input activation statistics. AWQ leverages the interaction between weights and activations to minimize output distortion. \textbf{SignRound} \cite{cheng2023optimize} introduces a trainable rounding adjustment parameter \( V \) and two weight clipping parameters \( \alpha, \beta \). Given calibration data \( \mathcal{D} \), the model \( M \), and block module \( m_w \) with weights \( w \), it iteratively updates \( V, \alpha, \) and \( \beta \) to minimize the reconstruction loss. The quantization function is defined as $\tilde{w} = qdq(w, \alpha, \beta, V)$ using a scaling factor $s = \frac{\max(W) * \alpha - \min(W) * \beta}{2^{{\text{bit}}} - 1}$ and rounding adjustment: $\tilde{W} = s \cdot \text{clip} \left( \frac{W}{s} + zp + V, n, m \right)$. The loss function \( mse(y_{quant}, y_{float}) \) minimizes the Frobenius norm $\|WX - \tilde{W}X\|_F^2$. 
Gradient updates leverage SignSGD, updating weights as $W_{t+1} = W_t - lr_t \cdot \text{sign}(g_t)$, allowing efficient navigation in constrained parameter spaces such as \( [-0.5, 0.5] \) for rounding and \( [0,1] \) for weight clipping.

 \begin{figure*}[t!]
 \centering
        \subfloat[MolmoE-1B]{
            \includegraphics[width=.25\linewidth]{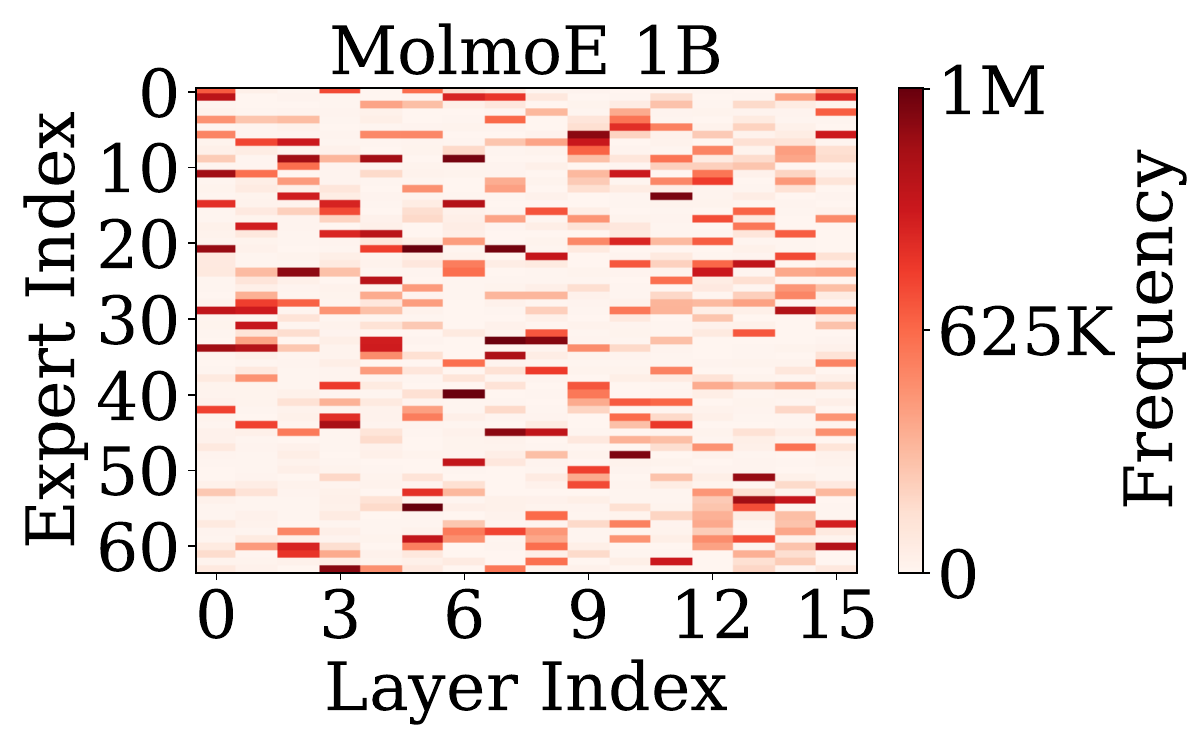}
            \label{subfig:molmoe_per_layer_freq}
        }
        \subfloat[DeepSeek VL2-Tiny]{
            \includegraphics[width=.25\linewidth]{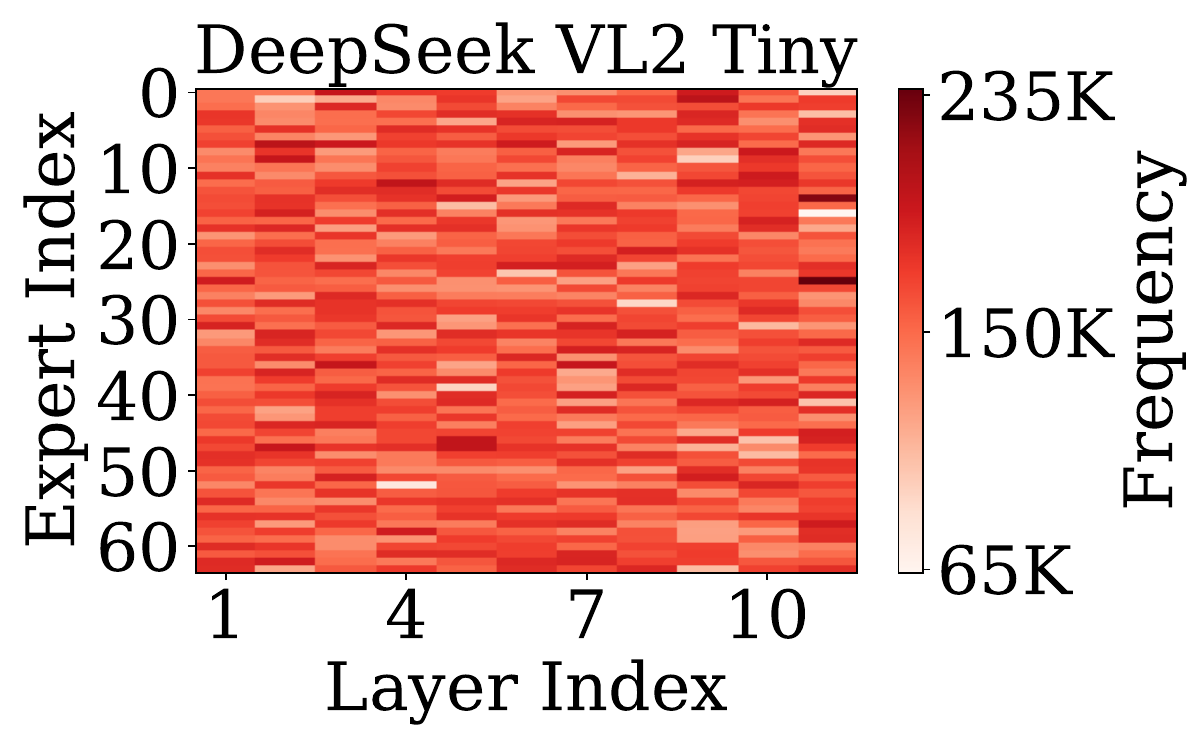}
            \label{subfig:deepseek_vl2_tiny_per_layer_freq}
        }
        \subfloat[DeepSeek VL2-Small]{
            \includegraphics[width=.25\linewidth]{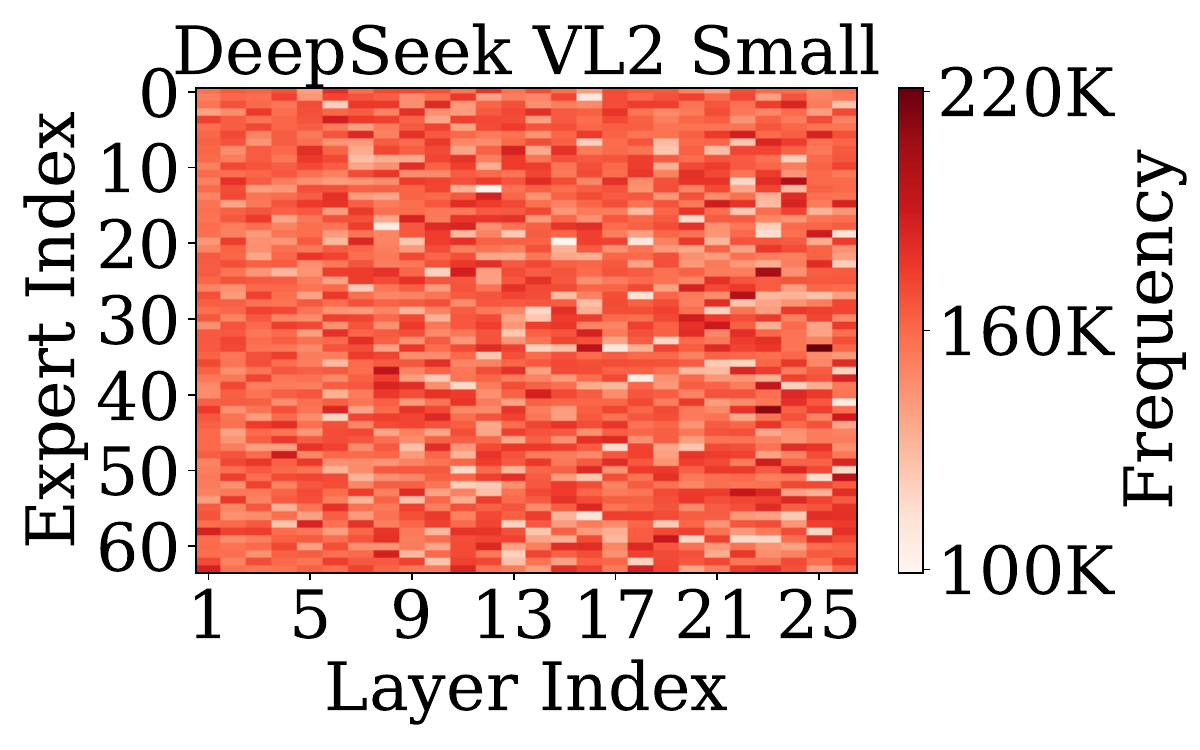}
            \label{subfig:deepseek_vl2_small_per_layer_freq}
        }
        \subfloat[DeepSeek VL2-Base]{
            \includegraphics[width=.25\linewidth]{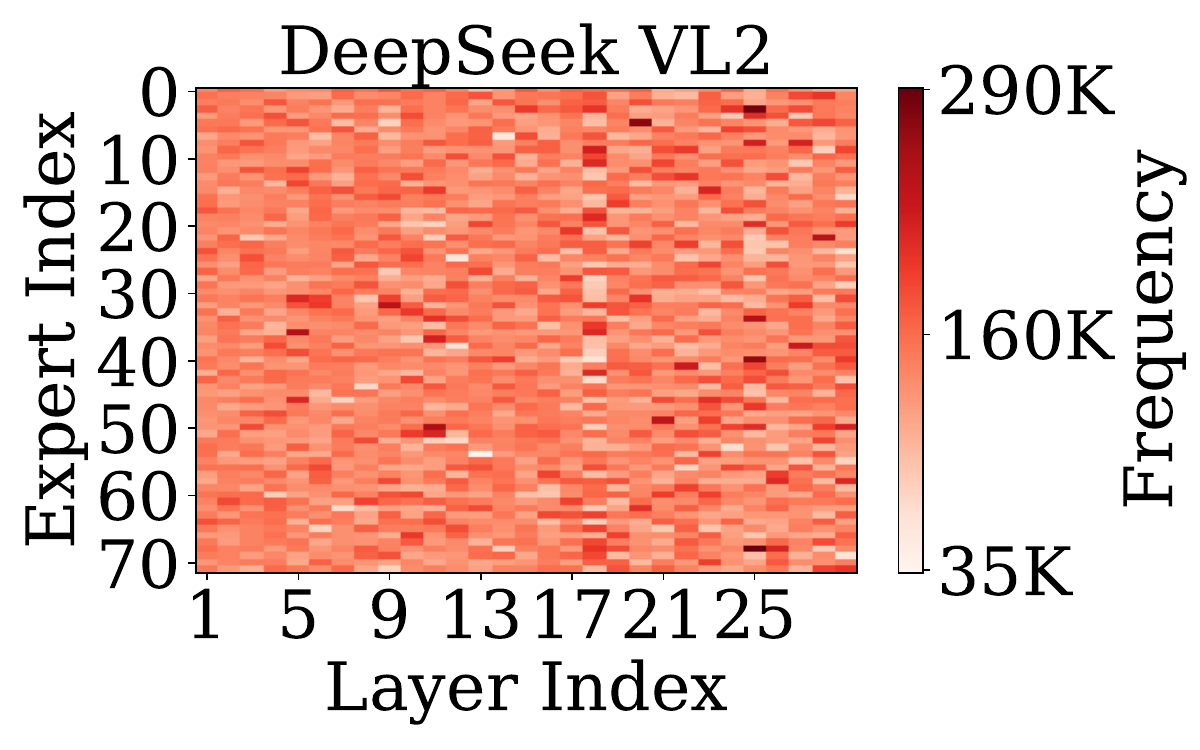}
            \label{subfig:deepseek_vl2_per_layer_freq}
        }
        \vspace{-2mm}
        \caption{Expert Activation Frequency map of MolmoE-1B and DeepSeek VL2 family Models on MME task}
        \vspace{-3mm}
        \label{fig:activation_freq}
    \end{figure*}

\begin{figure*}[t!]
 \centering
        \subfloat[MolmoE-1B]{
            \includegraphics[width=.25\linewidth]{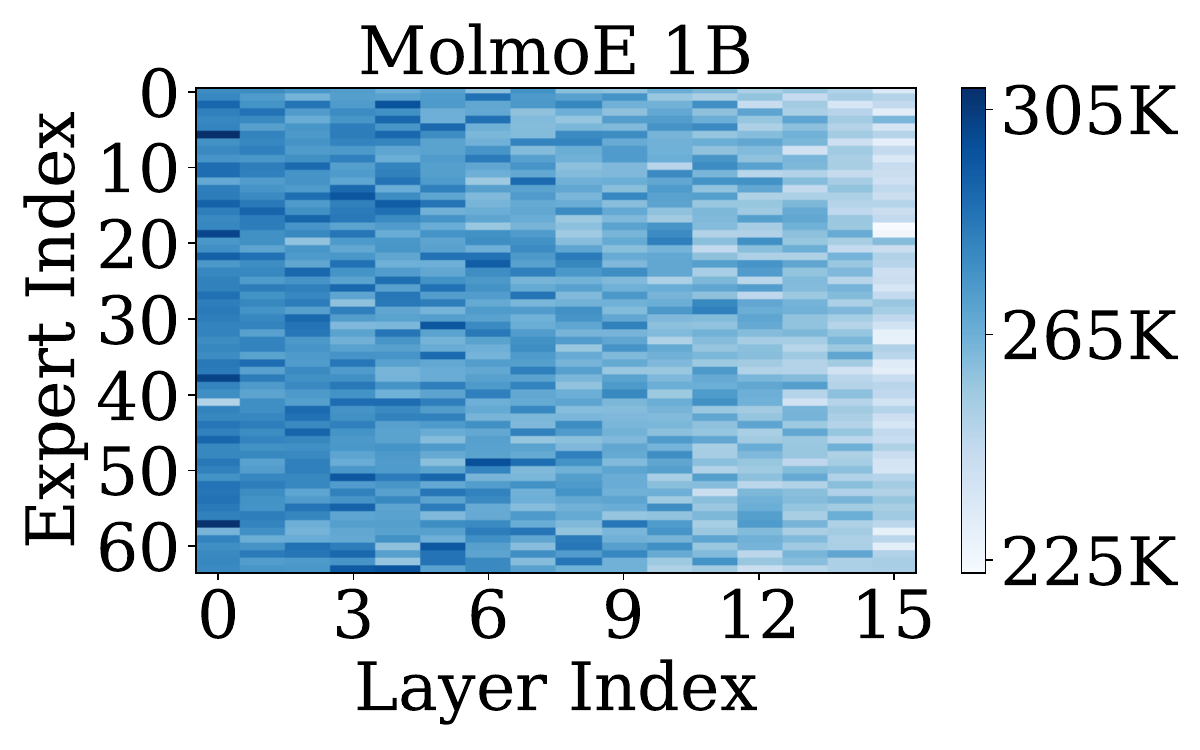}
            \label{subfig:molmoe_per_layer_hessian_trace}
        }
        \subfloat[DeepSeek VL2-Tiny]{
            \includegraphics[width=.25\linewidth]{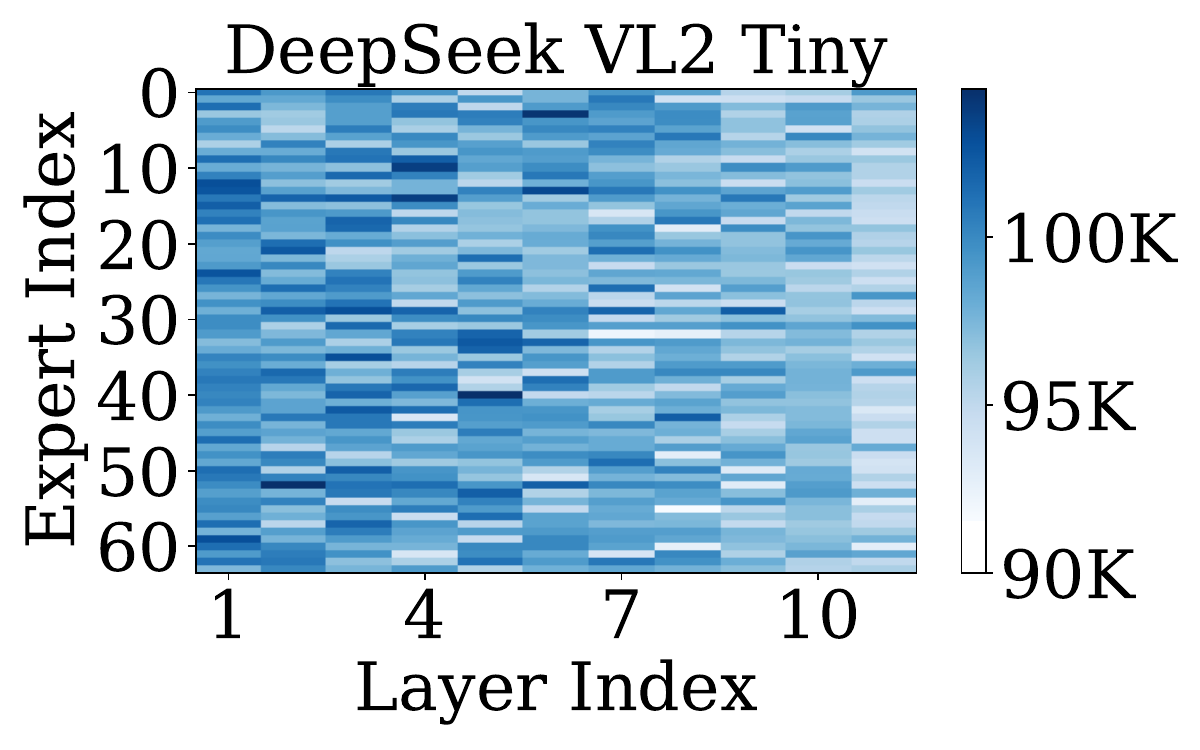}
            \label{subfig:deepseek_vl2_tiny_per_layer_hessian_trace}
        }
        \subfloat[DeepSeek VL2-Small]{
            \includegraphics[width=.25\linewidth]{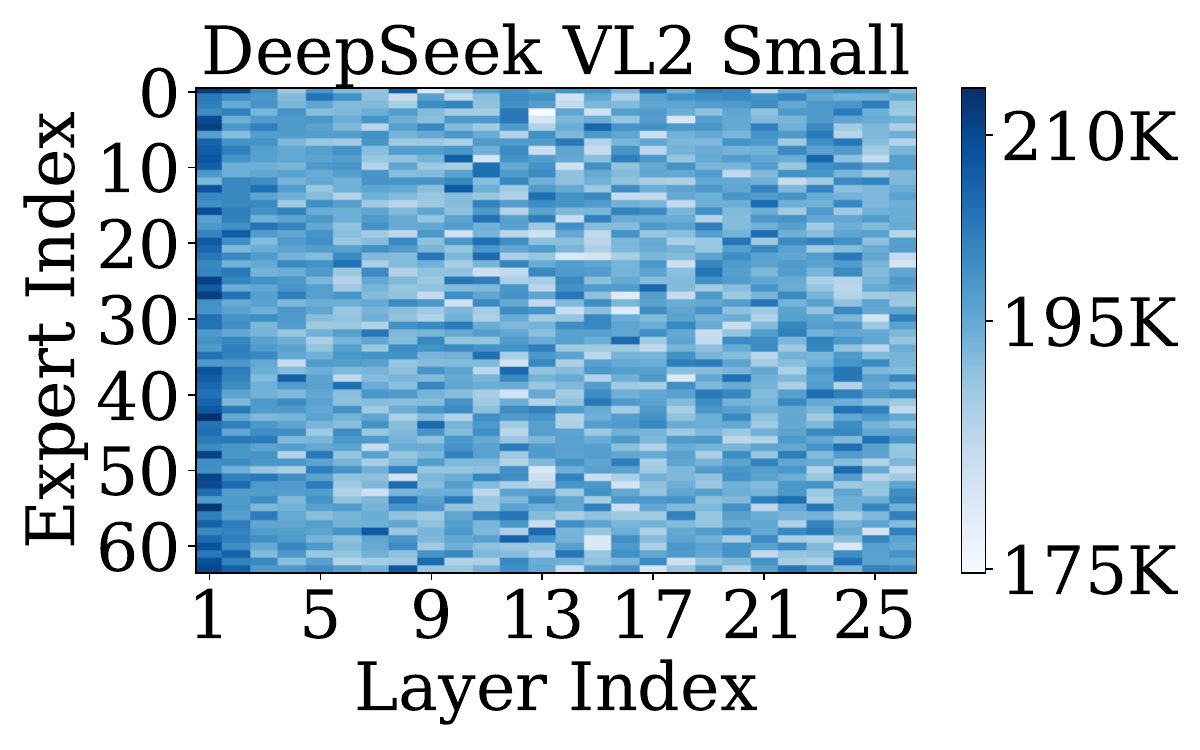}
            \label{subfig:deepseek_vl2_small_per_layer_hessian_trace}
        }
        \subfloat[DeepSeek VL2-Base]{
            \includegraphics[width=.25\linewidth]{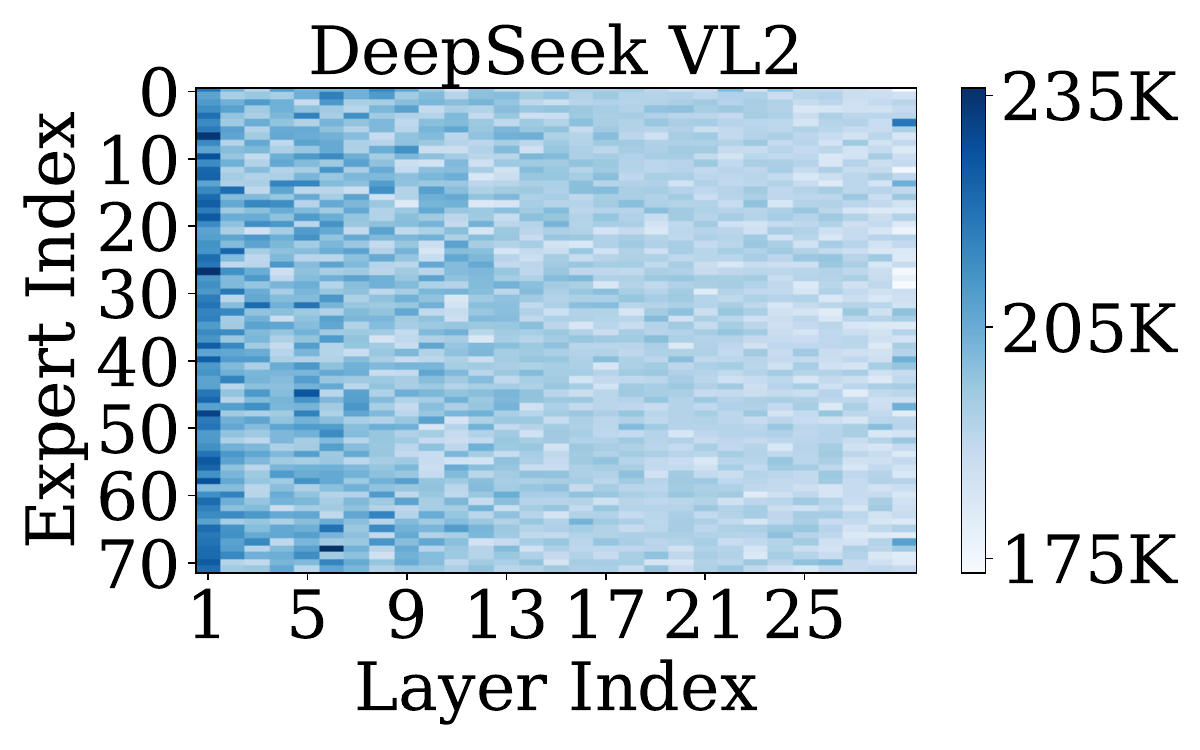}
            \label{subfig:deepseek_vl2_per_layer_hessian_trace}
        }
        \vspace{-2mm}
        \caption{Hessian Trace Approximation Map of MolmoE-1B and DeepSeek VL2 family Models}
        \vspace{-3mm}
        \label{fig:hessian_trace}
    \end{figure*}
\begin{figure*}[t!]
 \centering
        \subfloat[MolmoE-1B]{
            \includegraphics[width=.25\linewidth]{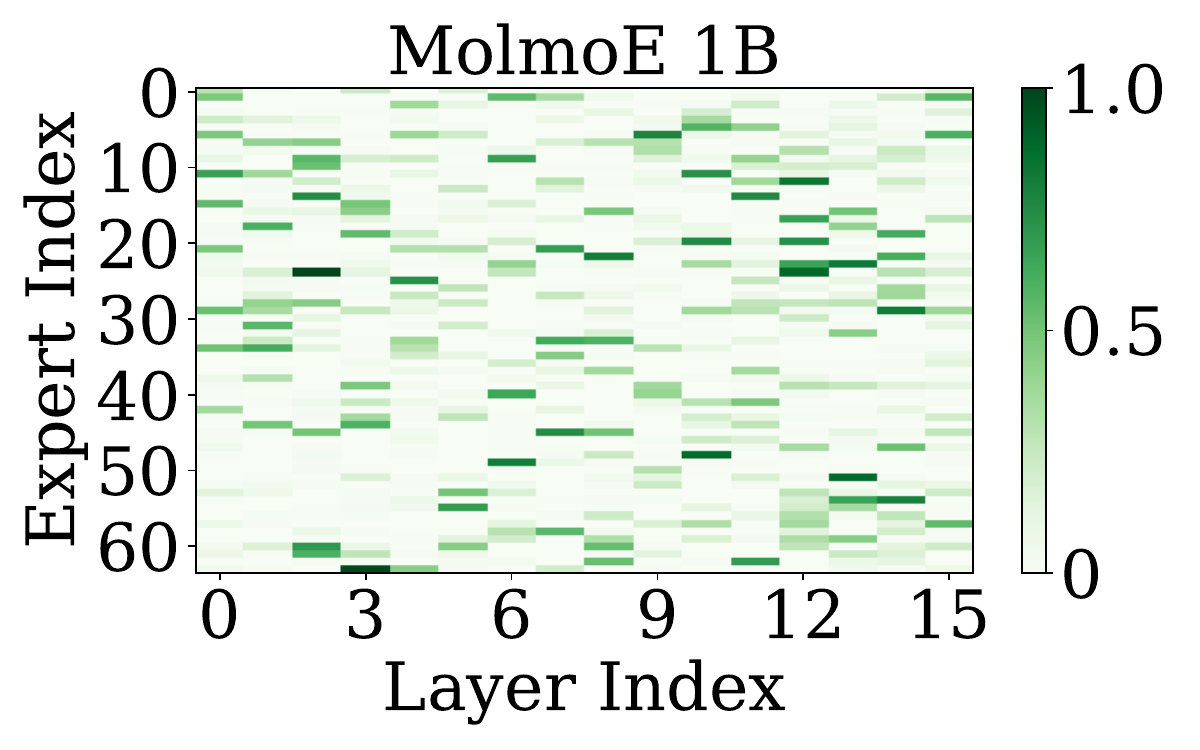}
            \label{subfig:molmoe_per_layer_combined_3_clusters}
        }
        \subfloat[DeepSeek VL2-Tiny]{
            \includegraphics[width=.25\linewidth]{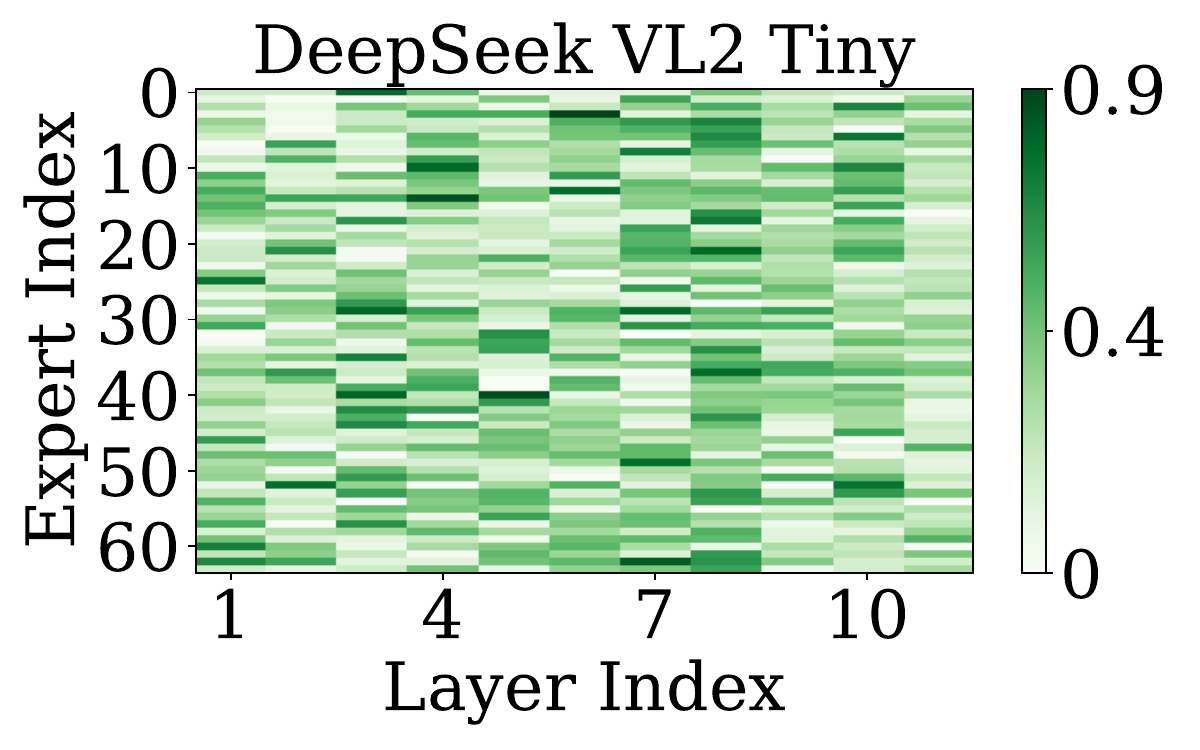}
            \label{subfig:deepseek_vl2_tiny_per_layer_combined_3_clusters}
        }
        \subfloat[DeepSeek VL2-Small]{
            \includegraphics[width=.25\linewidth]{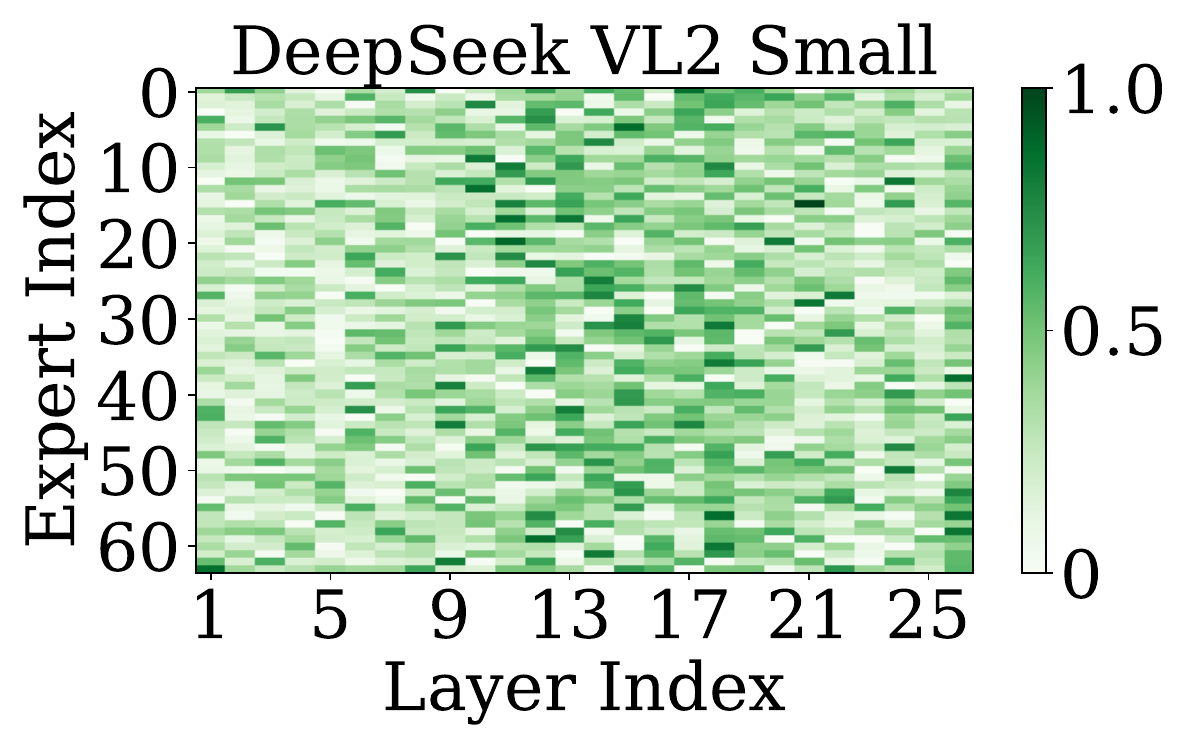}
            \label{subfig:deepseek_vl2_small_per_layer_combined_3_clusters}
        }
        \subfloat[DeepSeek VL2-Base]{
            \includegraphics[width=.25\linewidth]{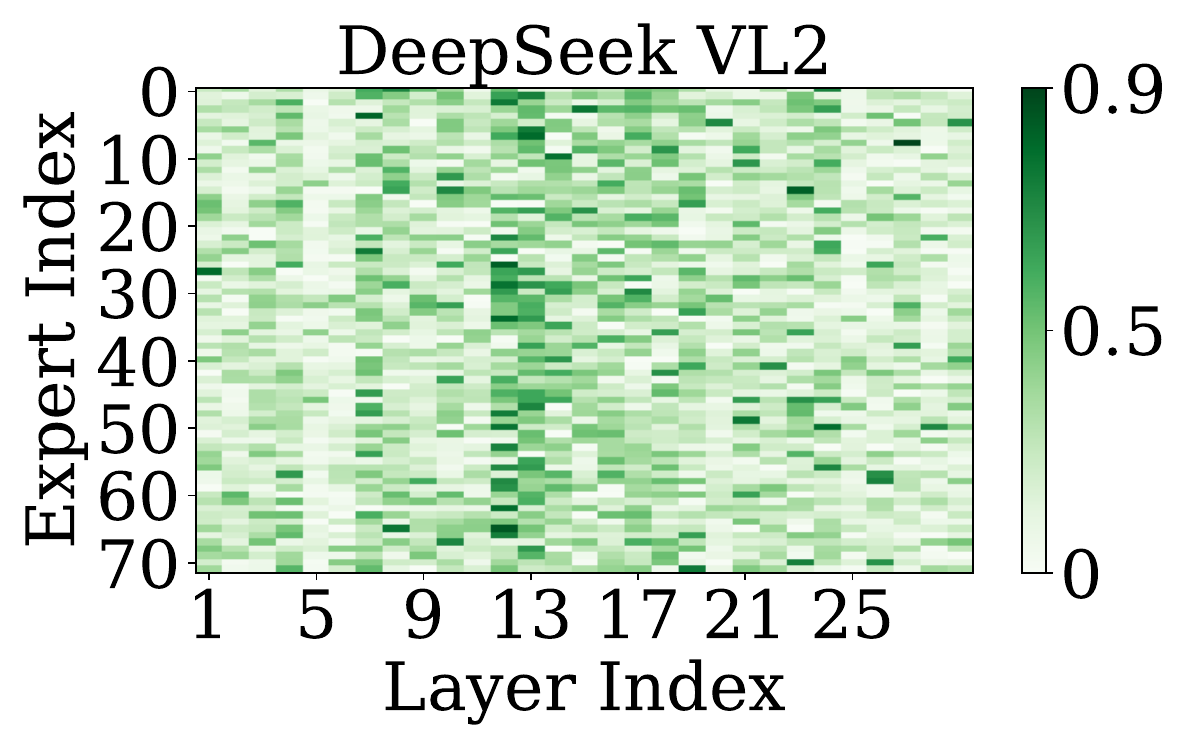}
            \label{subfig:deepseek_vl2_per_layer_combined_3_clusters}
        }
        \vspace{-2mm}
        \caption{Normalized Activation Frequency-Hessian Trace Approximation Importance Map}
        \vspace{-4mm}
        \label{fig:combined_3_clusters}
\end{figure*}



\section{Expert Profiling}\label{Profiling}

In this section, we analyze each expert's activation frequency using a calibration dataset and profile the sensitivity of expert using Hessian trace approximation.

\subsection{Vision Language Model Benchmarks}

We evaluate the performance of the following VLM-MoE models: DeepSeek VL2-Tiny \cite{deepseek_vl2-tiny}, DeepSeek VL2-Small \cite{deepseek_vl2-small}, DeepSeek VL2 \cite{deepseek_vl2-base}, and MolmoE-1B \cite{MolmoE_1B_0924}. DeepSeek-VL2 \cite{wu2024deepseek} models are built on top of DeepSeek-V2 \cite{liu2024deepseek}, while MolmoE-1B \cite{deitke2024molmo} is built on top of Olmo \cite{muennighoff2024olmoe}. The models vary in parameter count, layer depth, and expert configuration. The VLM benchmarks allow us to systematically compare the trade-offs between parameter efficiency, computational complexity, and expert utilization across varying scales of MoE architectures. The relevant configurations of these models are summarized in Table \ref{Tab:vlm_benchmarks}. The four VLM-MoE models have the MoE layer in the language component of the model. 
Note that DeepSeek-V2 model does not include MoE in the first transformer layer.

\begin{table}[H]
\centering
\caption{Summary of VLM-MoE Benchmarks. \textbf{P}: Number of Total Weight Parameters, \textbf{L}: Layers, \textbf{E}: Experts Per Layer, \textbf{AE}: Active Experts Per Token Per Layer}
\label{Tab:vlm_benchmarks}
\begin{tabular}{|c|c|c|c|c|}
\hline
Model              & \#P (B) & \#L & \#E & \#AE \\ \hline
DeepSeek VL2-Tiny  & 3       & 12  & 64  & 6    \\ \hline
DeepSeek VL2-Small & 16      & 27  & 64  & 6    \\ \hline
DeepSeek VL2       & 27      & 30  & 72  & 6    \\ \hline
MolmoE-1B          & 7.2     & 16  & 64  & 8    \\ \hline
\end{tabular}
\end{table}

\subsection{Expert Activation Frequency}

Figure \ref{fig:activation_freq} illustrates the expert activation frequency (number of times each expert is used in inference) heatmap for DeepSeek-VL2 family and MolmoE-1B models using the MME task dataset \cite{fu2023mme}. DeepSeek-VL2 family models exhibit a relatively uniform activation pattern across experts and layers, while MolmoE-1B demonstrates a more sparse activation pattern, with certain experts being activated more frequently. The expert activation frequency in MolmoE-1B is significantly higher, reaching up to 1M for certain experts, compared to DeepSeek-VL2 models, which peak at around 290K. This is because DeepSeek-V2 \cite{liu2024deepseek} employs an auxiliary loss during training to balance expert utilization, ensuring all experts are uniformly activated. Therefore, activation frequency alone is not a reliable indicator of expert importance in well-balanced models.


\subsection{Per-Expert Hessian Trace Approximation}

Traditionally, a calibration dataset is required to precisely compute the expert activation statistics. However, this may introduce bias and overlook sensitivity considerations, requiring data-free approaches. Therefore, we use Hessian trace approximation $H$ to compute the expert sensitivity. Hessian trace is defined as $H_{ij} = \frac{\partial^2 \mathcal{L}}{\partial \theta_i \partial \theta_j}$, a second-order behavior of the optimization landscape \cite{dong2019hawq, dong2020hawq}. However, for large language and vision models, $d$ can be very high and computing the full Hessian becomes computationally intractable with $\mathcal{O}(d^2)$ space complexity and $\mathcal{O}(d^3)$ time complexity for operations like eigen decomposition.

To address this, we estimate the Hessian trace $\text{Tr}(\mathbf{H})$ of FC layers in an expert using the Hutchinson algorithm \cite{hutchinson1989stochastic, avron2011randomized} (Algorithm \ref{alg:hutchinson_frobenius}). We use the Frobenius norm as a proxy for the loss function, and hence, we do not require any calibration dataset. The Hessian trace is estimated as $\text{Tr}(\mathbf{H}) \approx \frac{1}{m} \sum_{i=1}^{m} \mathcal{T}[i]$ over $m$ random samples where $\mathcal{T}[i]$ is the estimated Hessian trace of sample $i$. This method leverages the mathematical identity $\text{Tr}(H) = \mathbb{E}_{v \sim \mathcal{N}(0,I)}[v^T H v]$. For every sample, a random vector (randomly chosen from Rademacher distribution) $\mathbf{v} \sim \mathcal{N}(0,1)$ with the same shape as $\mathbf{W}$ is sampled. The Frobenius norm loss ($\mathcal{L} = \|\mathbf{W}\|_F = \sqrt{\sum_{i,j} W_{ij}^2}$) is used as a surrogate loss function due to its strong correlation with the Hessian’s spectral properties, ensuring a stable approximation. The first-order gradient $\mathbf{g}_1 = \nabla_{\mathbf{W}} \mathcal{L}$ is computed. Then, the Hessian-vector product (HVP) is obtained via $\text{HVP} = \nabla_{\mathbf{W}} (\mathbf{g}_1^\top \mathbf{v})$ which avoids explicit Hessian computation. The trace estimated for each sample is given by $\mathcal{T}[i] = \sum (\mathbf{v} \odot \text{HVP})$. The final Hessian trace is obtained by averaging over all samples. Frobenius norm as a proxy loss is motivated by its smooth and differentiable nature, ensuring efficient gradient computation and stability in second-order approximations. Additionally, the Frobenius norm aligns with the Hessian’s structure, making it an effective approximation method for quantifying layer importance. 
The method is summarized in Algorithm \ref{alg:hutchinson_frobenius}. 
The Hessian trace of Expert $i$ is the sum of Hessian of Gate, Up and Down layers ($H_{i}^{G}$ + $H_{i}^{D}$ + $H_{i}^{U}$).

\begin{algorithm}
\caption{Hessian Trace Approximation using Hutchinson's Method for an FC Layer}
\label{alg:hutchinson_frobenius}
\begin{algorithmic}[1]
\Require Weight Tensor $\mathbf{W}$ 
\Require \#Samples $m$
\Ensure Estimated trace of the Hessian $\mathrm{Tr}(\mathbf{H})$
\State Initialize trace estimates array $\mathcal{T} \gets []$
\For{$i = 1$ to $m$}
    \State Sample $\mathbf{v} \sim \mathcal{N}(0, 1)$ 
    \State Proxy loss: $\mathcal{L} = \|\mathbf{W}\|_F$ (Frobenius norm)
    \State First-order gradient: $\mathbf{g}_1 = \nabla_{\mathbf{W}} \mathcal{L}$ with tracking
    \State Hessian-vector product (HVP): $\text{HVP} = \nabla_{\mathbf{W}} (\mathbf{g}_1^\top \mathbf{v})$
    \State Trace estimate: $\mathcal{T}[i] \gets \sum (\mathbf{v} \odot \text{HVP})$
\EndFor
\State Return Average Hessian Trace 
\State $\mathrm{Tr}(\mathbf{H}) = \frac{1}{m} \sum_{i=1}^{m} \mathcal{T}[i]$
\end{algorithmic}
\end{algorithm}

Figure \ref{fig:hessian_trace} illustrates the Hessian trace approximation heatmaps for MoE experts across the four VLM-MoE models.
The experts in deeper layers exhibit lower Hessian values compared to those in the initial layers, suggesting that the first layers are highly sensitive to quantization. DeepSeek VL2-Small demonstrates a more uniform trace distribution across its layers and experts than other models, possibly due to smoother gradients. Consequently, quantizing models like DeepSeek VL2-Small, which have uniform sensitivity across all layers and uniform expert activation, can be highly challenging.

\subsection{Normalized Activation Frequency-Hessian Trace Approximation}     
The expert activation frequency and Hessian sensitivity distribution differ across models. A highly sensitive expert may not be activated frequently, while a frequently activated expert may exhibit only moderate sensitivity. This variation arises due to the dynamic routing mechanism in MoE models, where expert selection is influenced by input tokens rather than solely by model sensitivity. Therefore, we combine both statistics to analyze their impact on model performance. With the combined approach, the overall importance ($I$) of expert $i$ is determined by the product of the normalized activation frequency ($AF$) and Hessian ($H$) sensitivity, as shown in the equation below.
\[
I_i = \left( \frac{AF_i - \min_j AF_j}{\max_j AF_j - \min_j AF_j} \right) 
\cdot \left( \frac{H_i - \min_j H_j}{\max_j H_j - \min_j H_j} \right)
\]

Figure \ref{fig:combined_3_clusters} illustrates the normalized activation frequency-hessian sensitivity map of different models. In the MolmoE-1B model, only a small subset of experts still consistently exhibit high importance, reflecting uneven utilization. DeepSeek VL2-Tiny and Base models are more important in the middle layers than in the last and initial layers.


\begin{figure*}[t!]
 \centering
        \subfloat[MolmoE-1B]{
            \includegraphics[width=.24\linewidth]{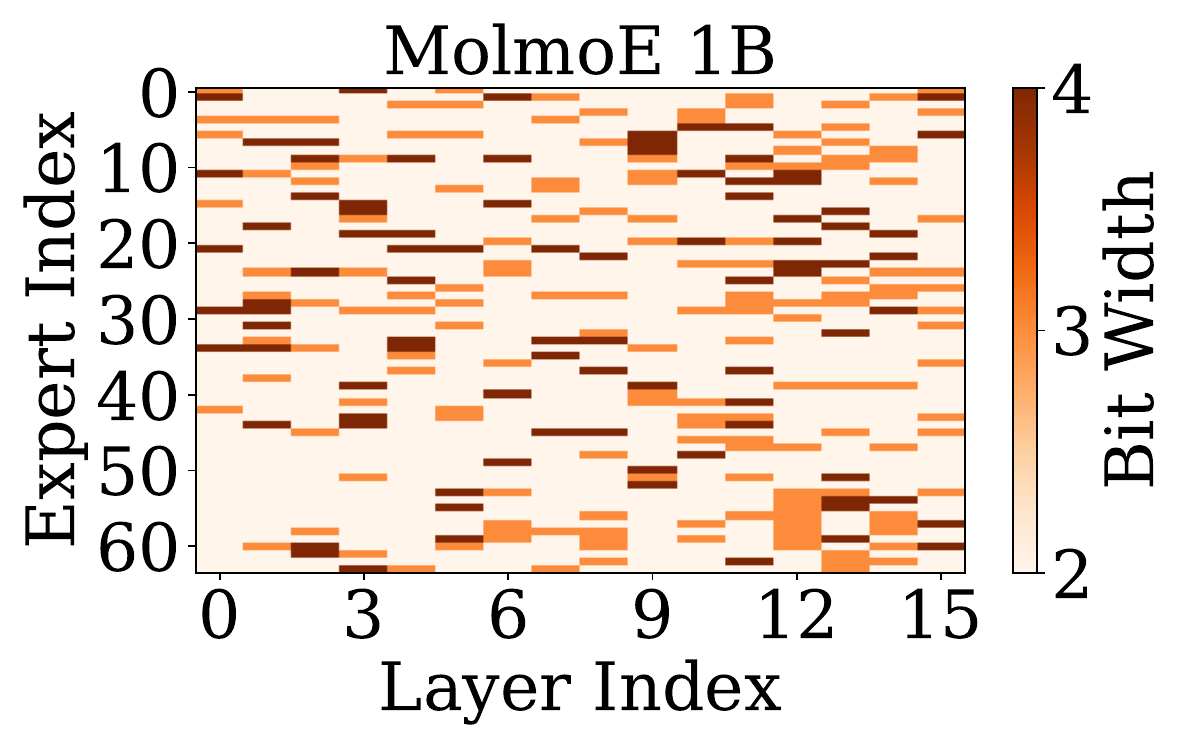}
            \label{subfig:molmoe_vl2_per_layer_freq_3_clusters}
        }
        \subfloat[DeepSeek VL2-Tiny]{
            \includegraphics[width=.24\linewidth]{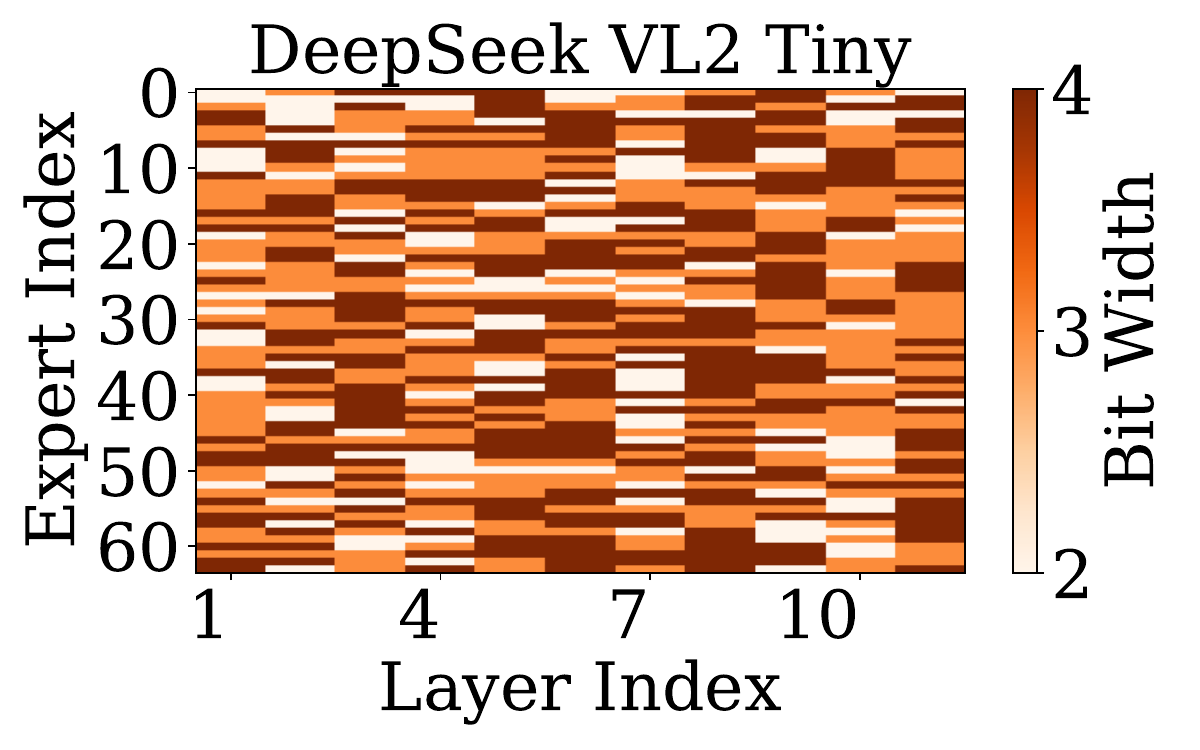}
            \label{subfig:ds_vl2_tiny_per_layer_freq_3_clusters}
        }
        \subfloat[DeepSeek VL2-Small]{
            \includegraphics[width=.24\linewidth]{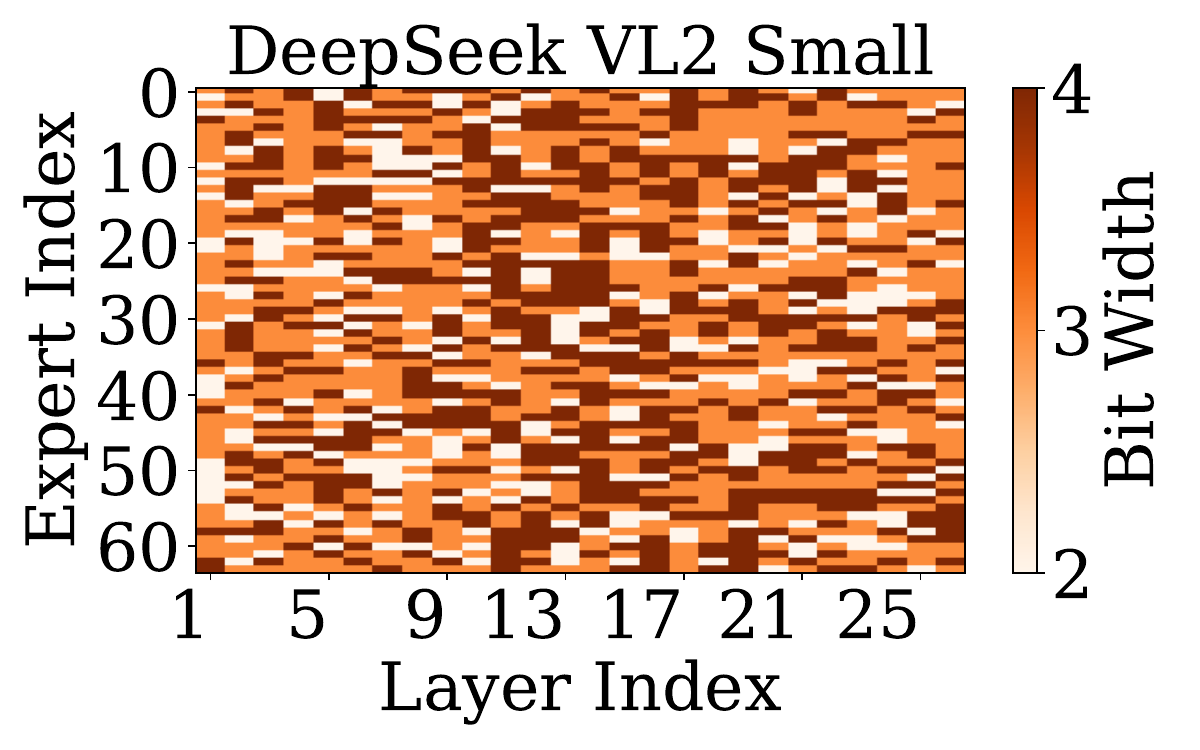}
            \label{subfig:ds_vl2_small_per_layer_freq_3_clusters}
        }
        \subfloat[DeepSeek VL2-Base]{
            \includegraphics[width=.24\linewidth]{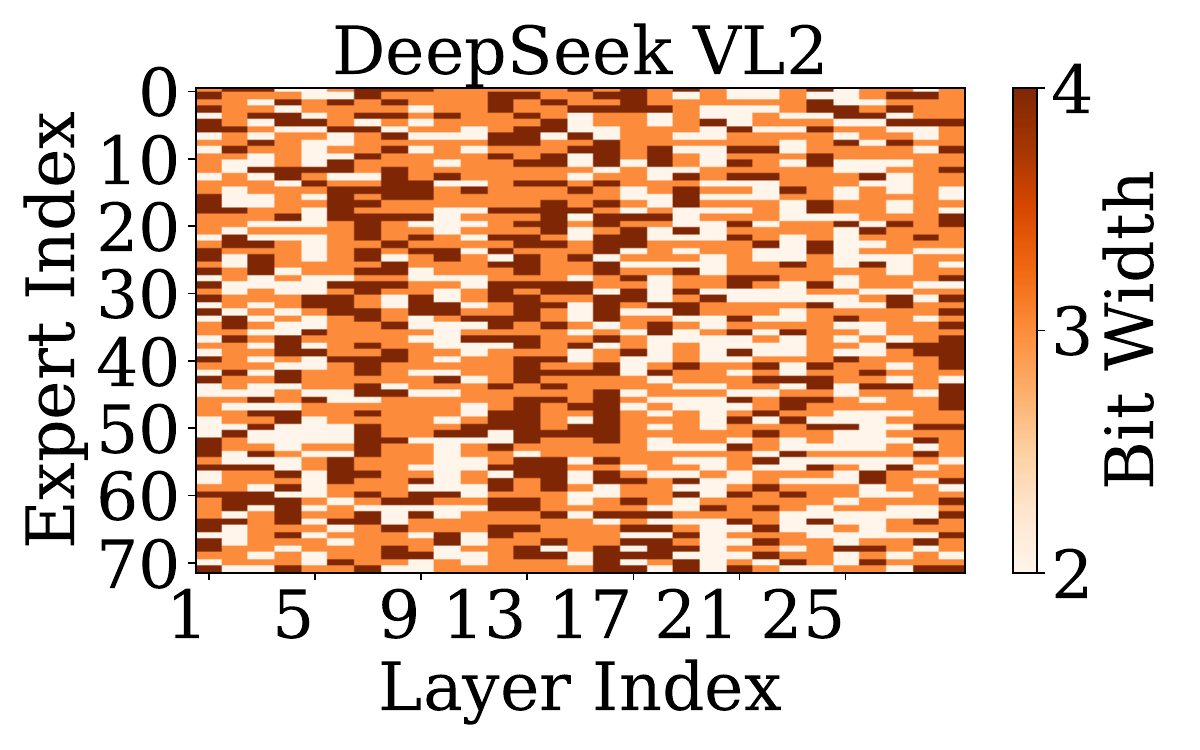}
            \label{subfig:ds_vl2_per_layer_freq_3_clusters}
        }
        \vspace{-2mm}
        \vspace{-0.8mm}
        \caption{Layer-wise Precision Assignment Map based on Expert Activation Frequency}
        \vspace{-3mm}
        \label{fig:per_layer_freq_3_clusters}
\end{figure*}
\begin{figure*}[t!]
 \centering
        \subfloat[MolmoE-1B]{
            \includegraphics[width=.24\linewidth]{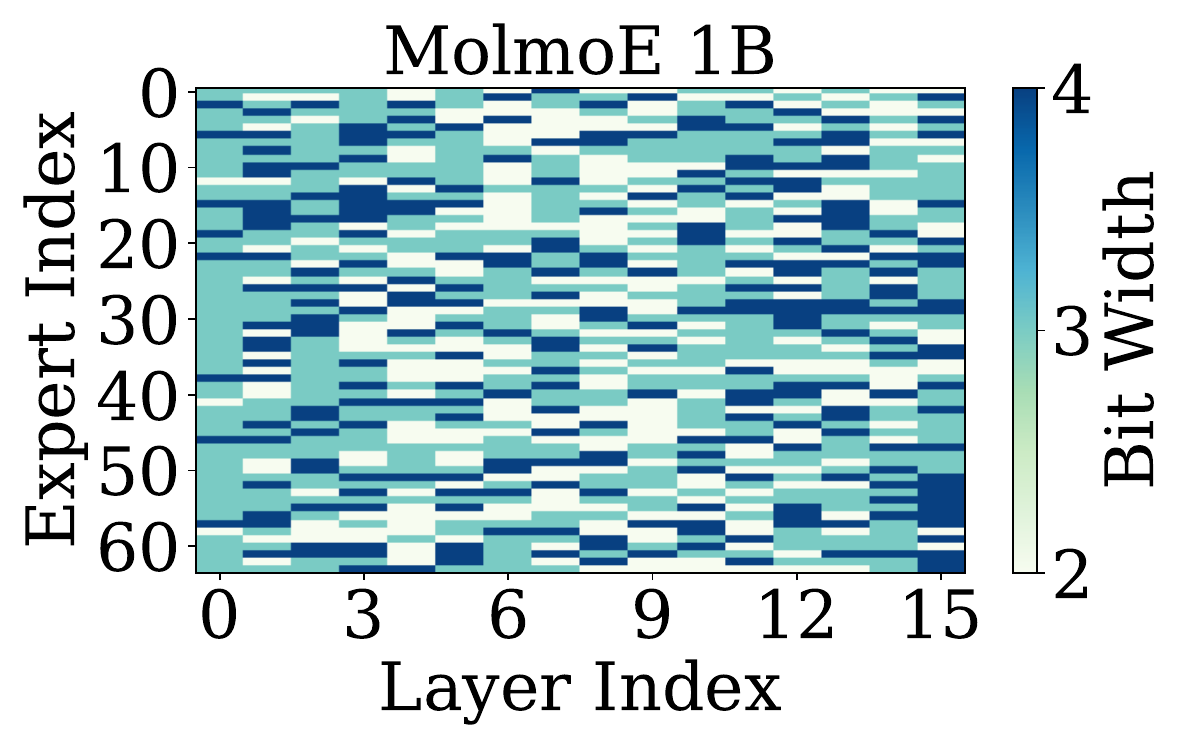}
            \label{subfig:molmoe_vl2_per_layer_hessian_3_clusters}
        }
        \subfloat[DeepSeek VL2-Tiny]{
            \includegraphics[width=.24\linewidth]{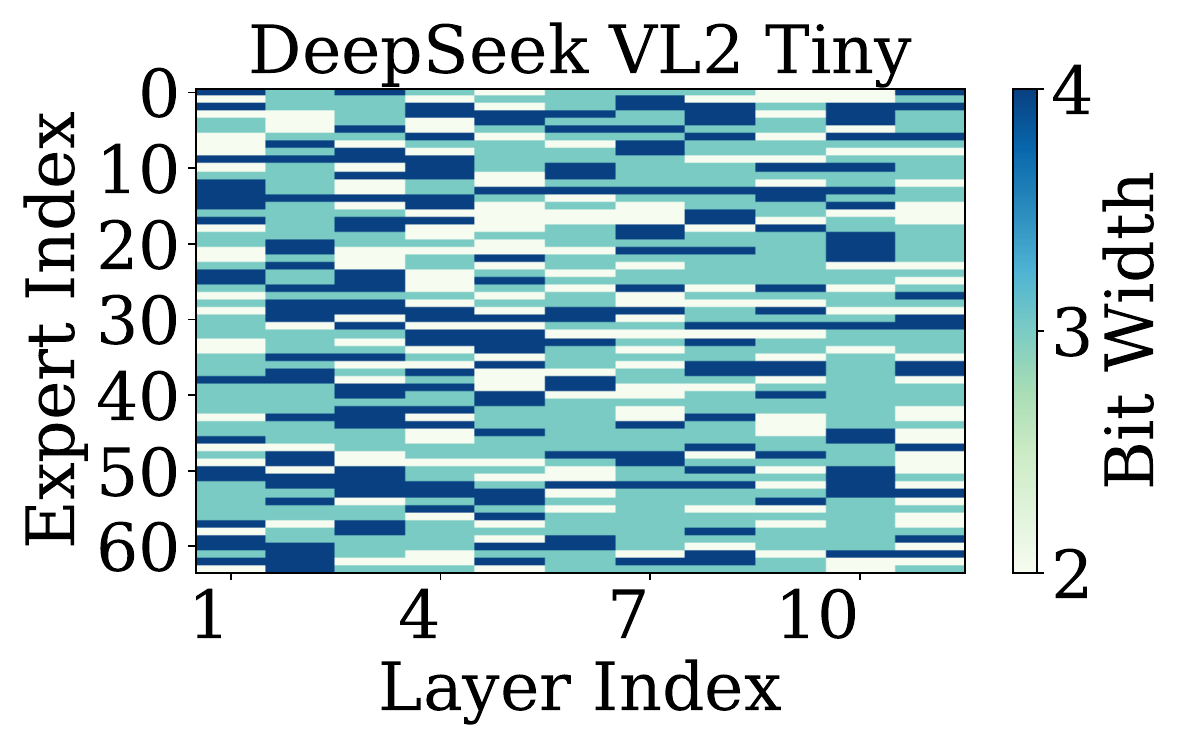}
            \label{subfig:ds_vl2_tiny_per_layer_hessian_3_clusters}
        }
        \subfloat[DeepSeek VL2-Small]{
            \includegraphics[width=.24\linewidth]{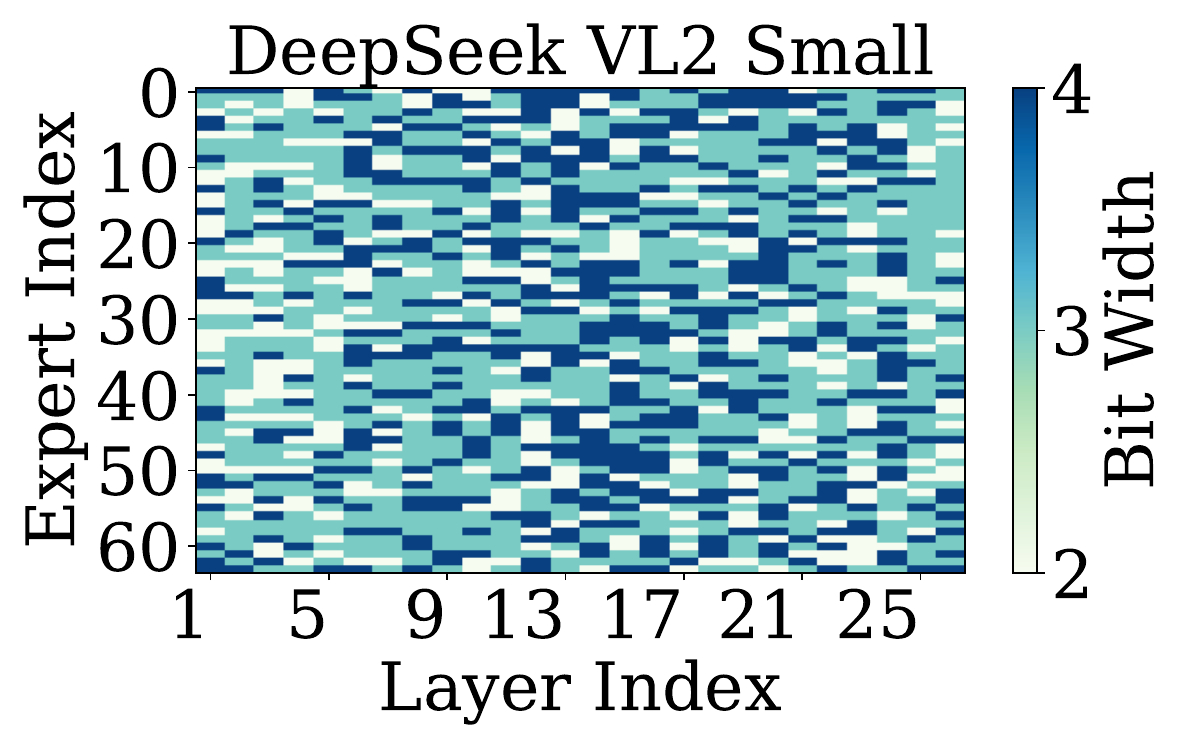}
            \label{subfig:ds_vl2_small_per_layer_hessian_3_clusters}
        }
        \subfloat[DeepSeek VL2-Base]{
            \includegraphics[width=.24\linewidth]{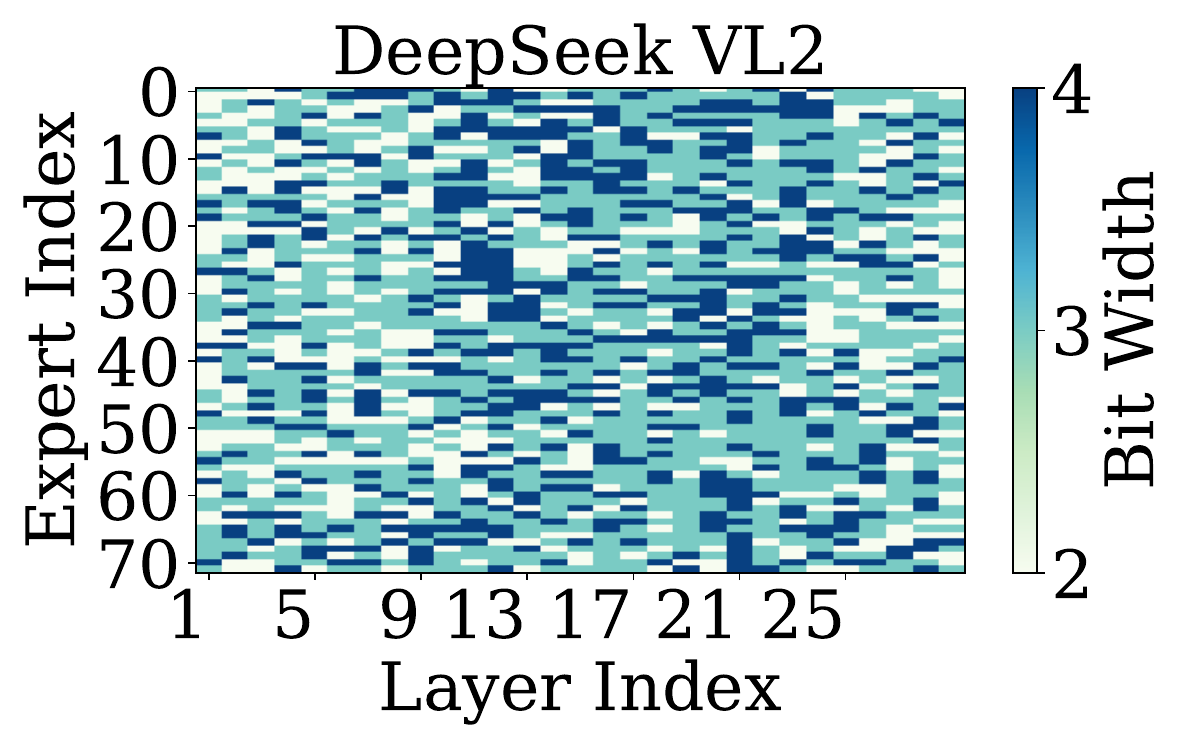}
            \label{subfig:ds_vl2_per_layer_hessian_3_clusters}
        }
        \vspace{-2mm}
        \caption{Layer-wise Precision Assignment Map based Hessian Trace Approximation}
        \vspace{-3mm}
        \label{fig:per_layer_hessian_3_clusters}
\end{figure*}

\begin{figure*}[t!]
 \centering
        \subfloat[MolmoE-1B]{
            \includegraphics[width=.24\linewidth]{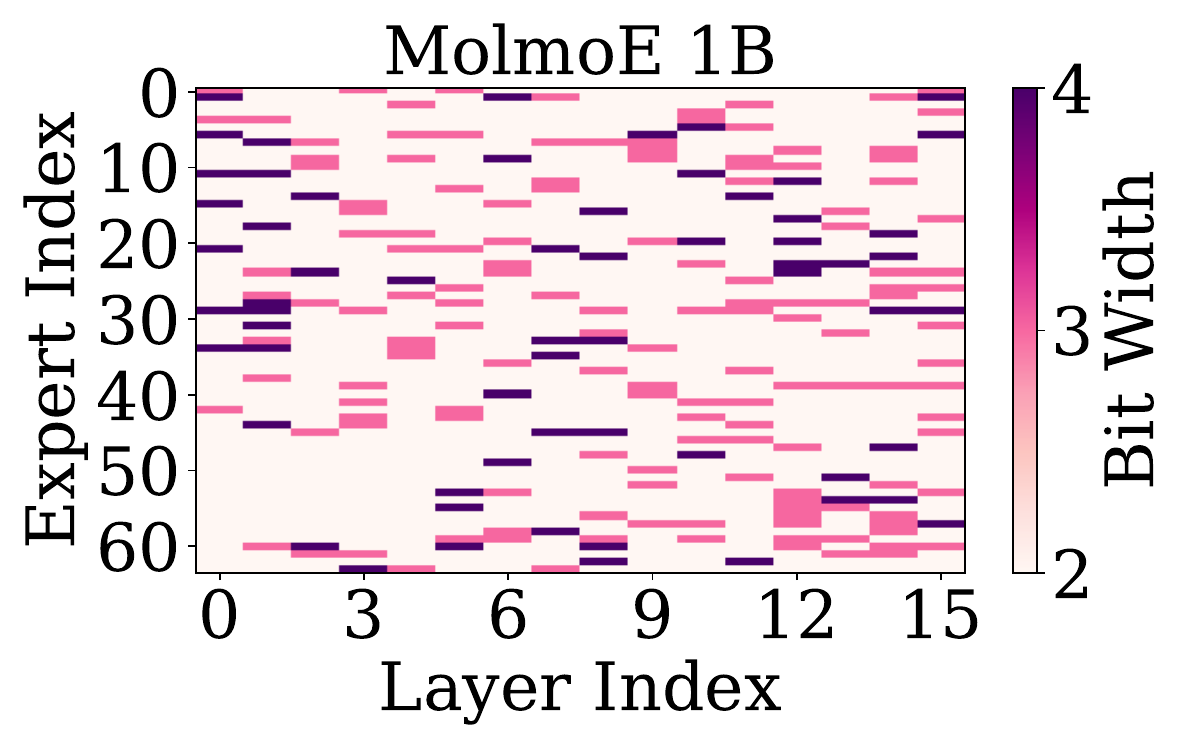}
            \label{subfig:molmoe_vl2_per_layer_combined_3_clusters}
        }
        \subfloat[DeepSeek VL2-Tiny]{
            \includegraphics[width=.24\linewidth]{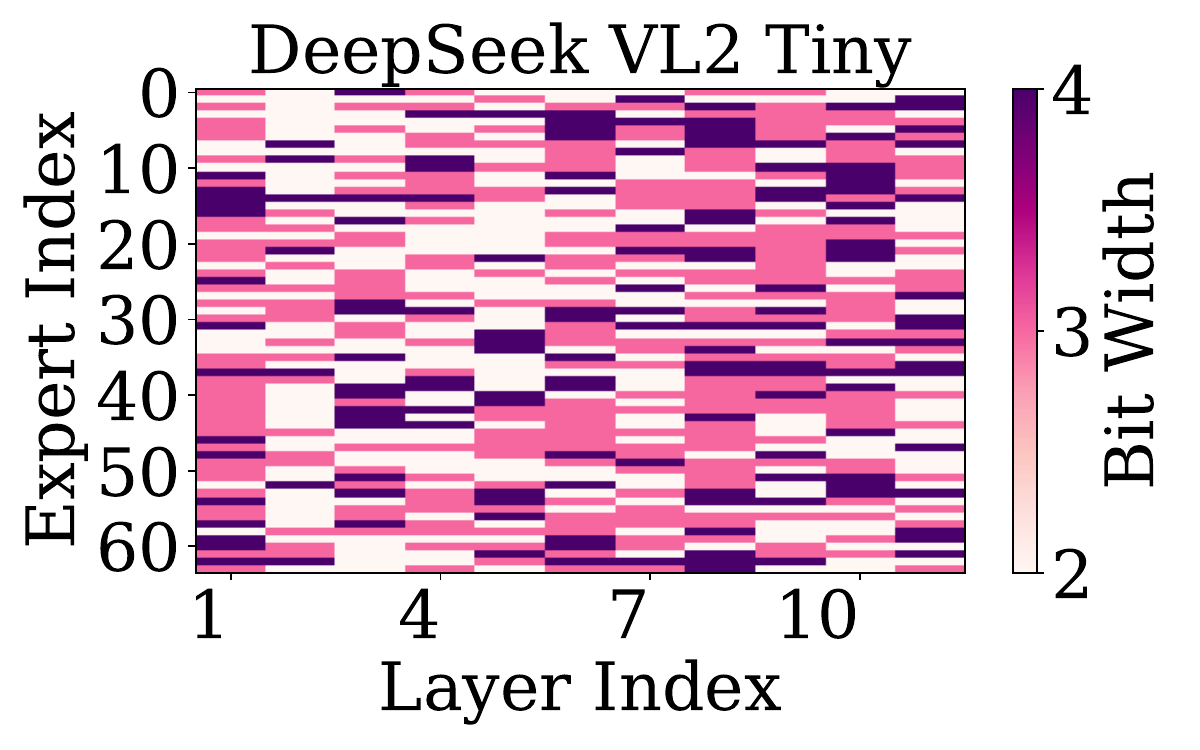}
            \label{subfig:ds_vl2_tiny_per_layer_combined_3_clusters}
        }
        \subfloat[DeepSeek VL2-Small]{
            \includegraphics[width=.24\linewidth]{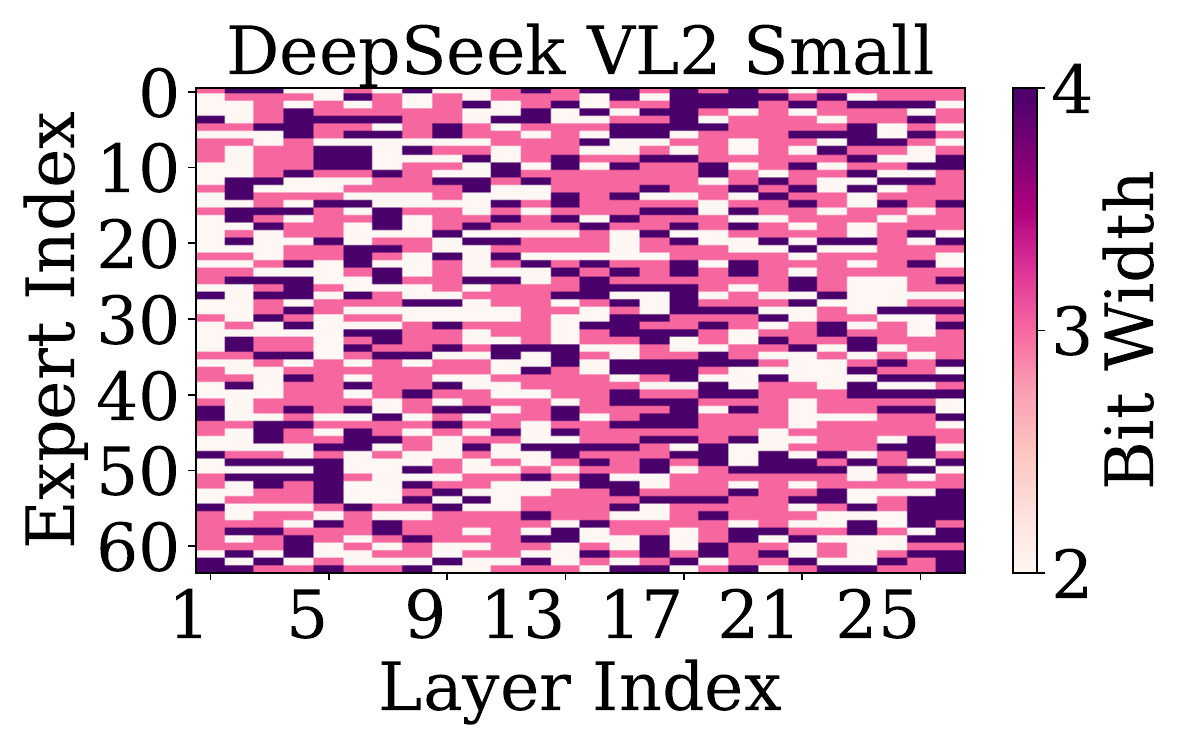}
            \label{subfig:ds_vl2_small_per_layer_combined_3_clusters}
        }
        \subfloat[DeepSeek VL2-Base]{
            \includegraphics[width=.24\linewidth]{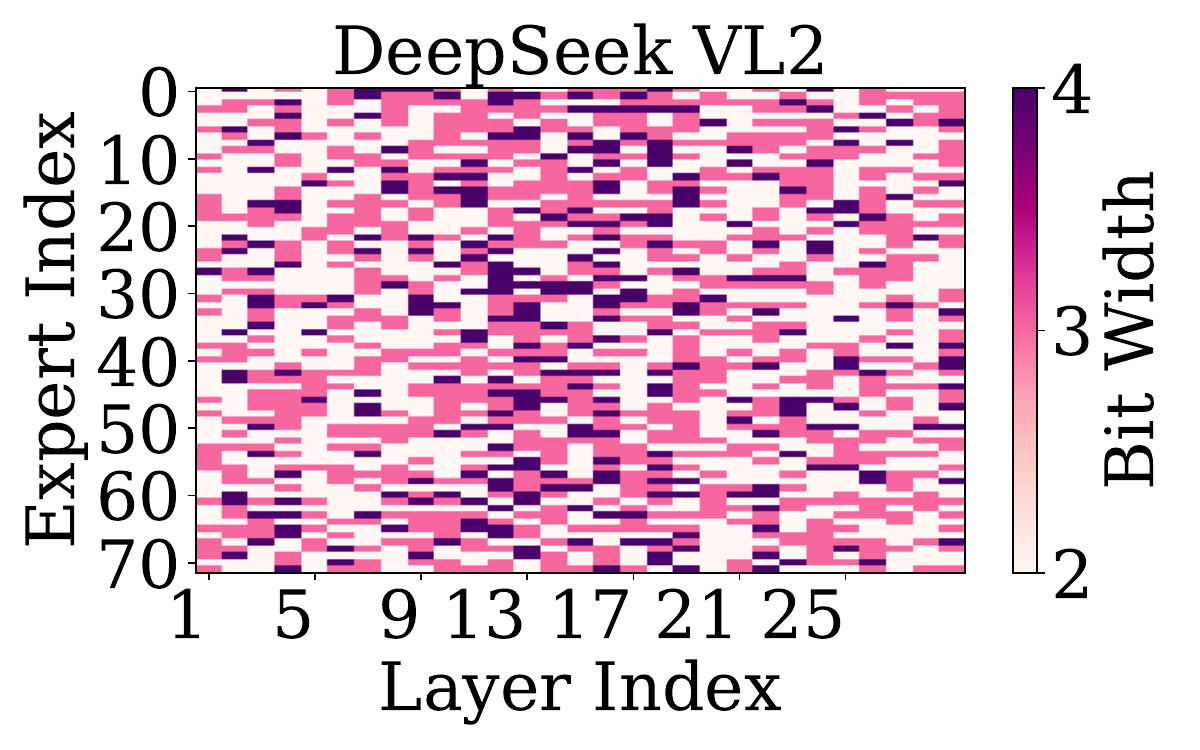}
            \label{subfig:ds_vl2_per_layer_combined_3_clusters}
        }
        \caption{Layer-wise Precision Assignment Map based on Normalized Hessian Trace \& Expert Activation Frequency}
        \label{fig:per_layer_combined_3_clusters}
\end{figure*}

\begin{figure*}[t!]
 \centering
        \subfloat[MolmoE-1B]{
            \includegraphics[width=.24\linewidth]{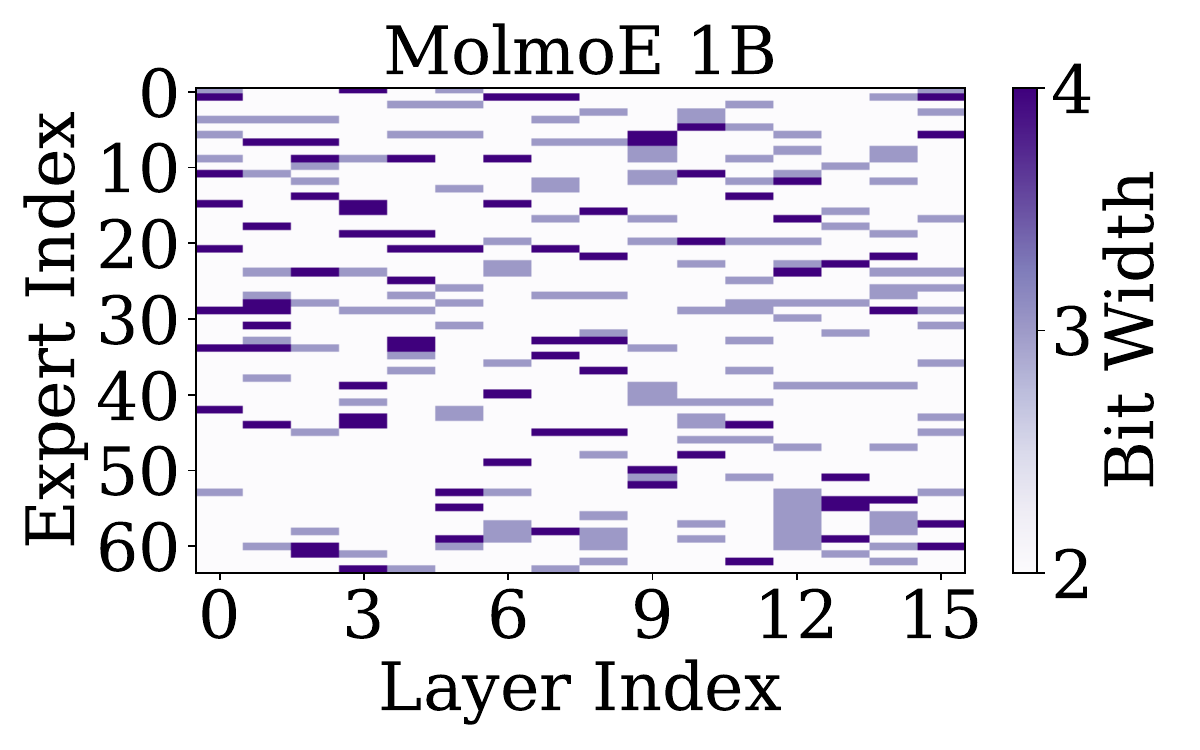}
            \label{subfig:molmoe_vl2_per_model_freq_3_clusters}
        }
        \subfloat[DeepSeek VL2-Tiny]{
            \includegraphics[width=.24\linewidth]{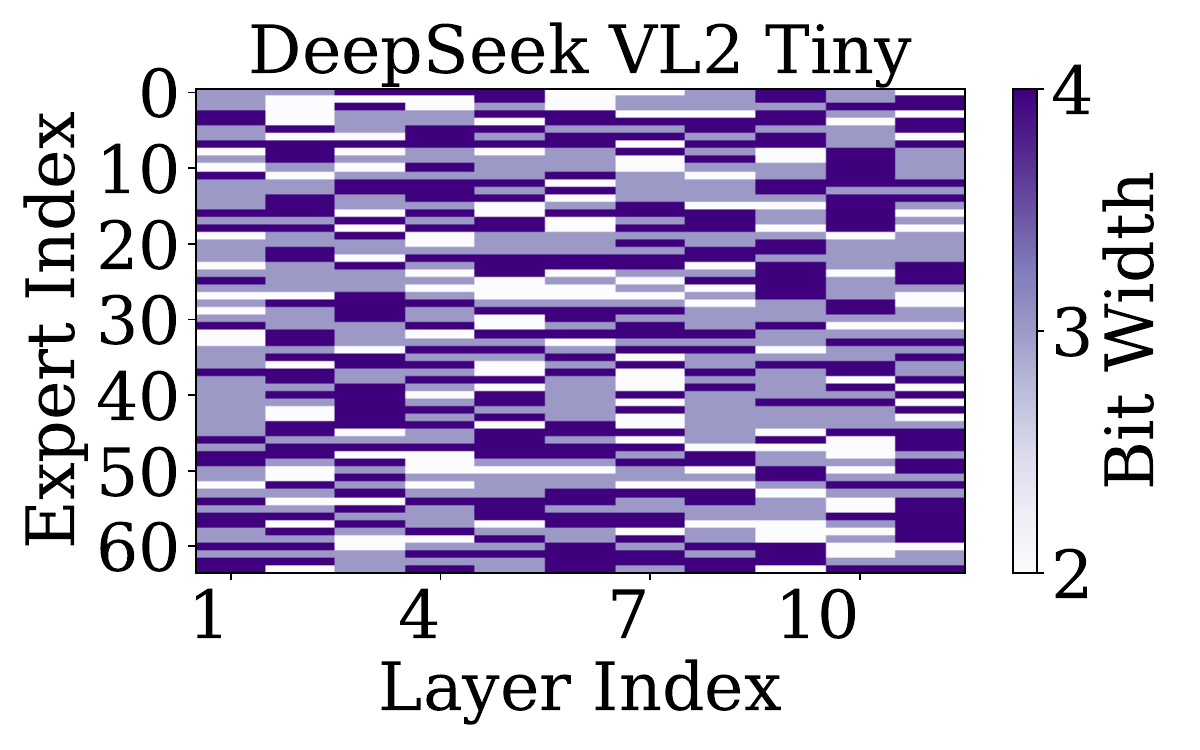}
            \label{subfig:ds_vl2_tiny_per_model_freq_3_clusters}
        }
        \subfloat[DeepSeek VL2-Small]{
            \includegraphics[width=.24\linewidth]{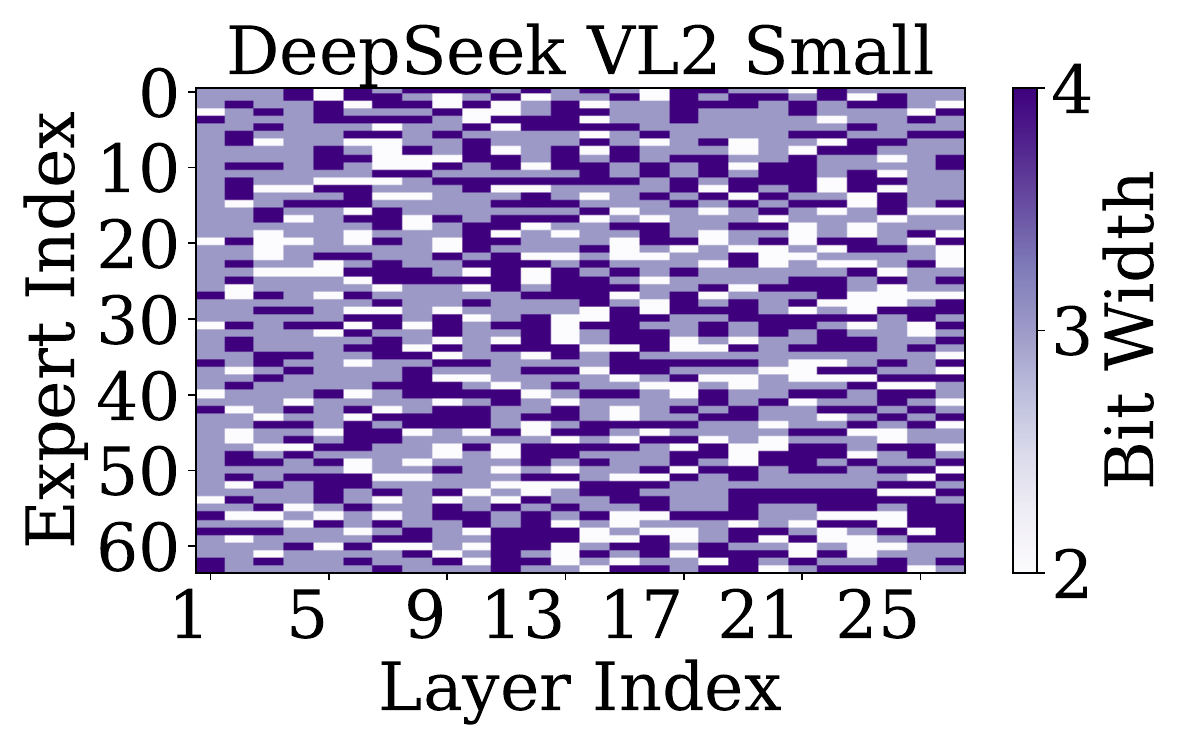}
            \label{subfig:ds_vl2_small_per_model_freq_3_clusters}
        }
        \subfloat[DeepSeek VL2-Base]{
            \includegraphics[width=.24\linewidth]{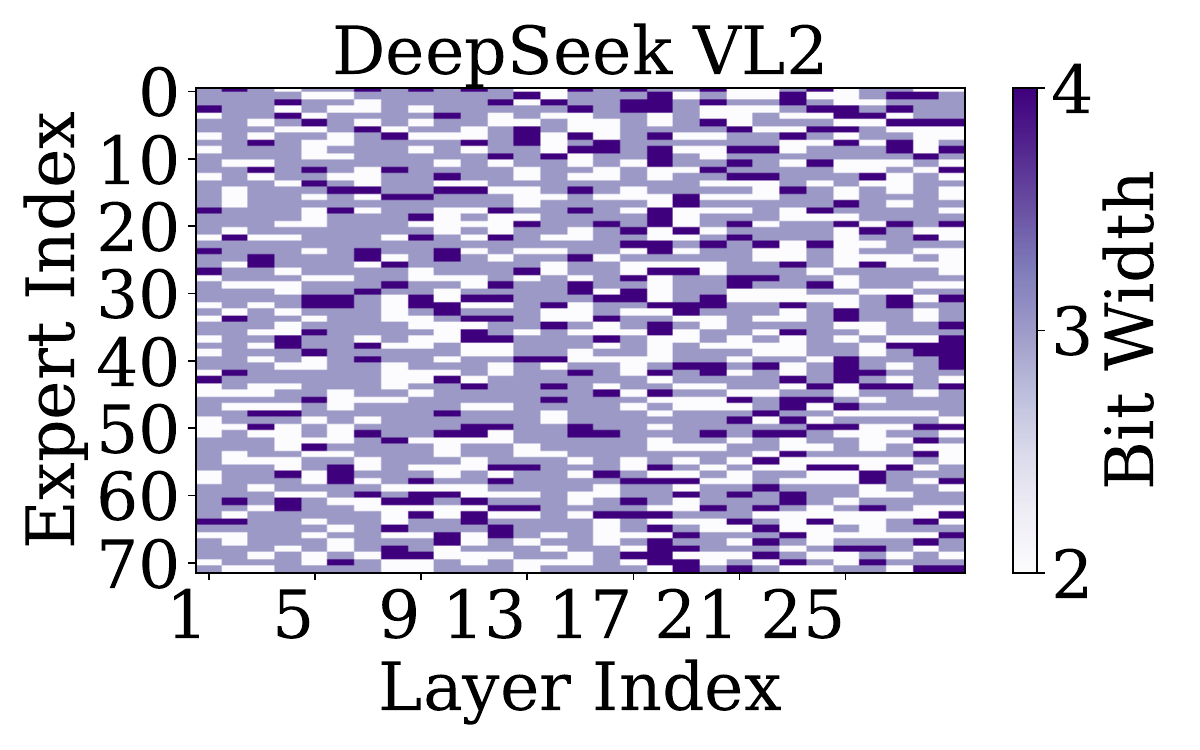}
            \label{subfig:ds_vl2_per_model_freq_3_clusters}
        }
        \vspace{-2mm}
        \caption{Model-wise Precision Assignment Map based on Expert Activation Frequency}
        \vspace{-3mm}
        \label{fig:per_model_freq_3_clusters}
\end{figure*}

\begin{figure*}[t!]
 \centering
        \subfloat[MolmoE-1B]{
            \includegraphics[width=.24\linewidth]{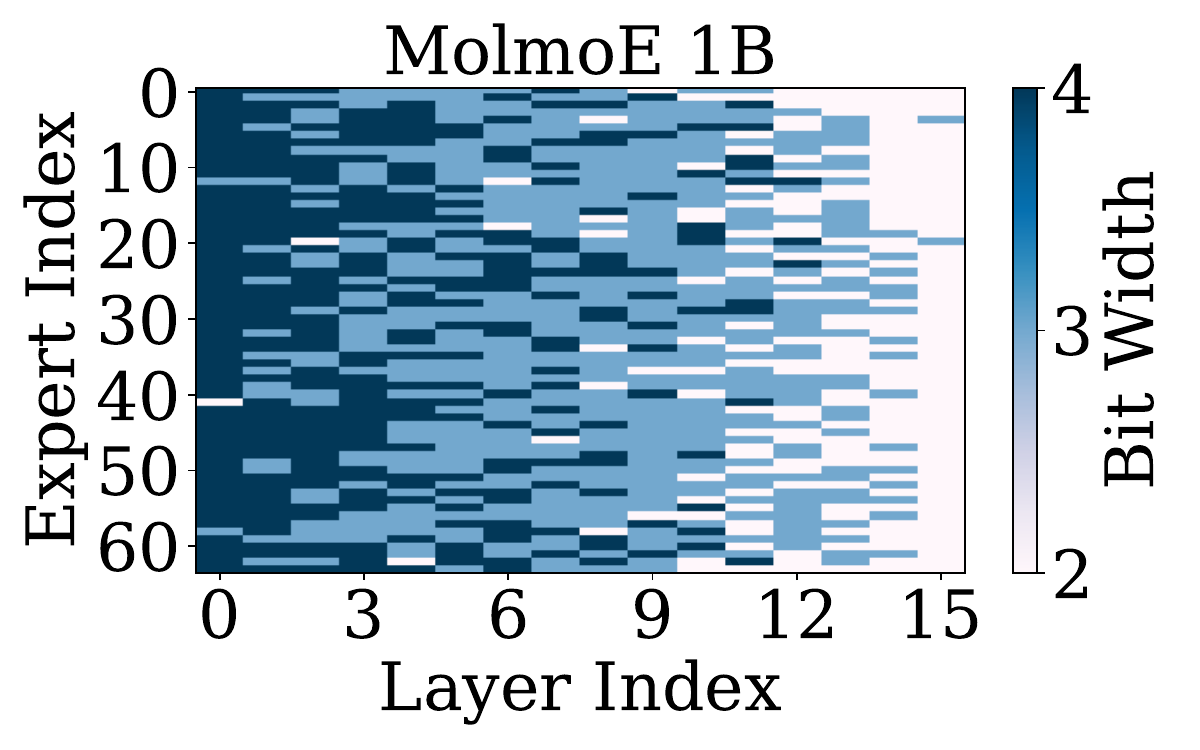}
            \label{subfig:molmoe_vl2_per_model_hessian_3_clusters}
        }
        \subfloat[DeepSeek VL2-Tiny]{
            \includegraphics[width=.24\linewidth]{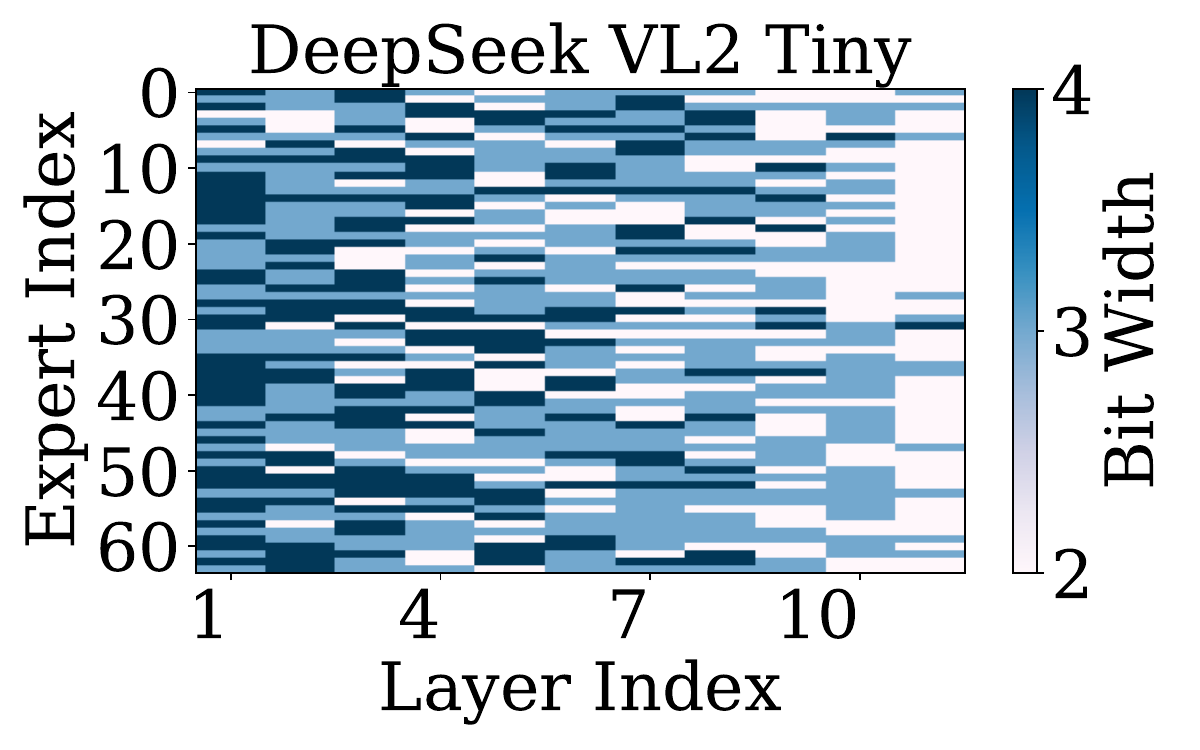}
            \label{subfig:ds_vl2_tiny_per_model_hessian_3_clusters}
        }
        \subfloat[DeepSeek VL2-Small]{
            \includegraphics[width=.24\linewidth]{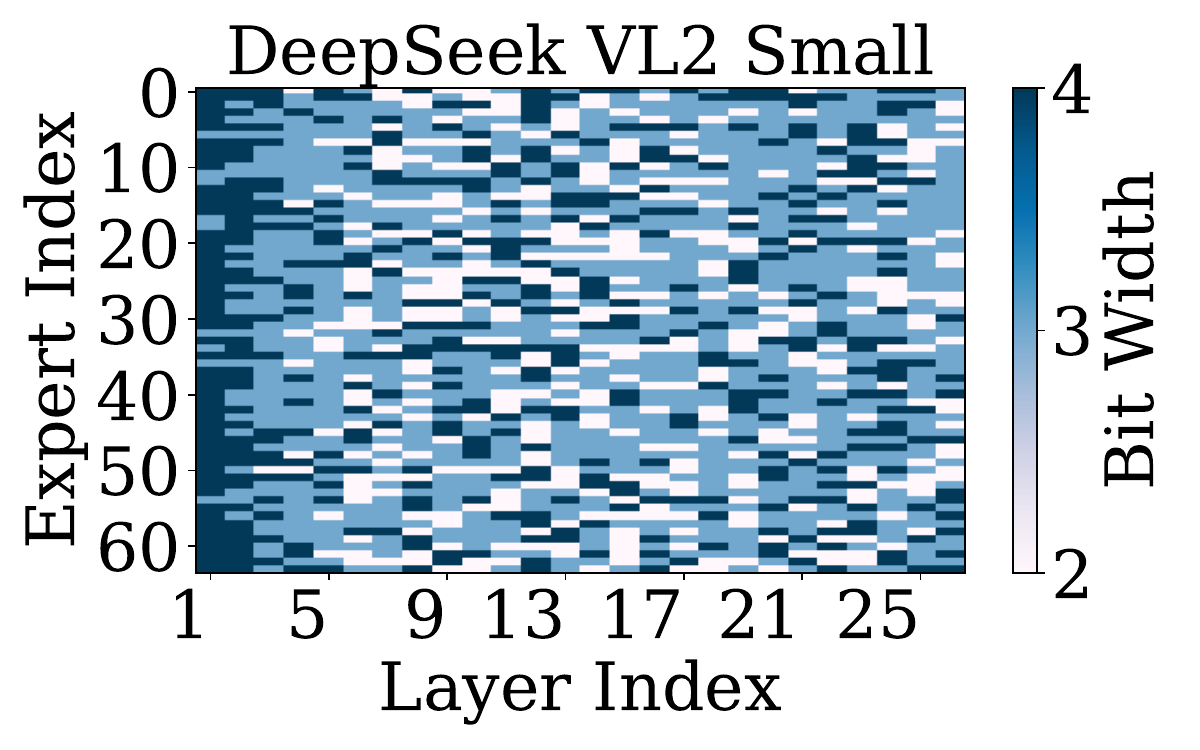}
            \label{subfig:ds_vl2_small_per_model_hessian_3_clusters}
        }
        \subfloat[DeepSeek VL2-Base]{
            \includegraphics[width=.24\linewidth]{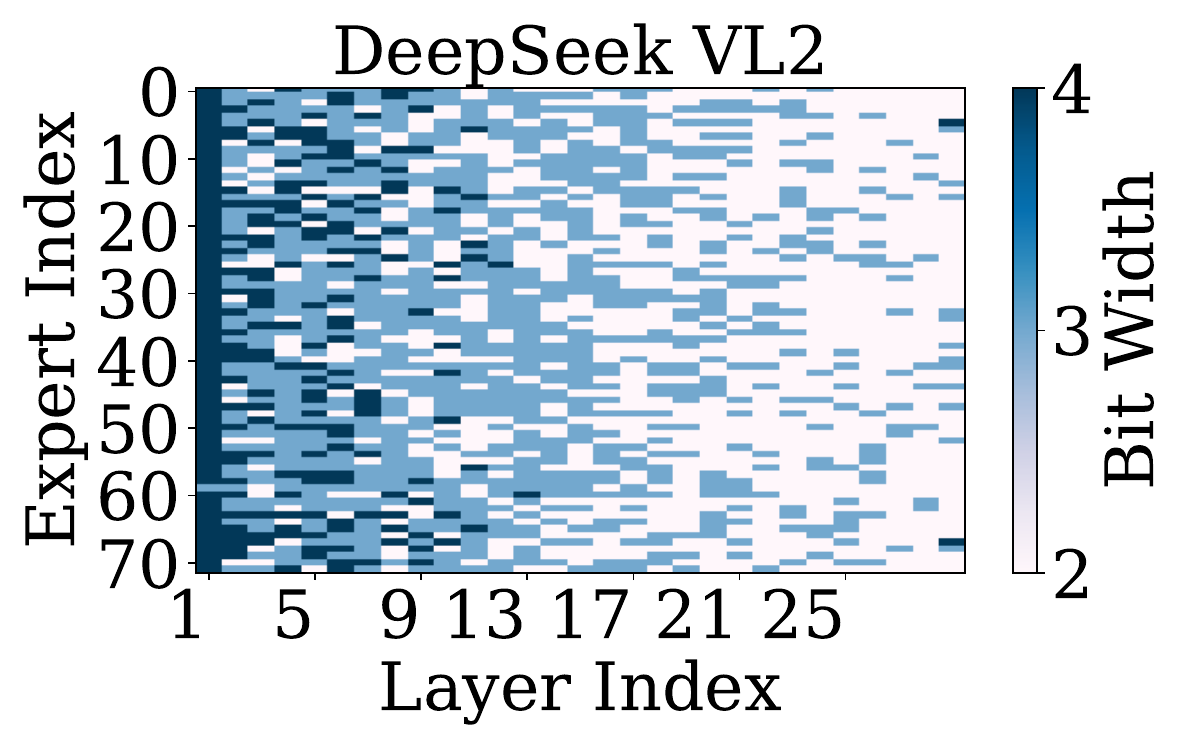}
            \label{subfig:ds_vl2_per_model_hessian_3_clusters}
        }
        \vspace{-2mm}
        \caption{Model-wise Precision Assignment Map based on Hessian Trace Approximation}
        \vspace{-3mm}
        \label{fig:per_model_hessian_3_clusters}
\end{figure*}

\begin{figure*}[t!]
 \centering
        \subfloat[MolmoE-1B]{
            \includegraphics[width=.24\linewidth]{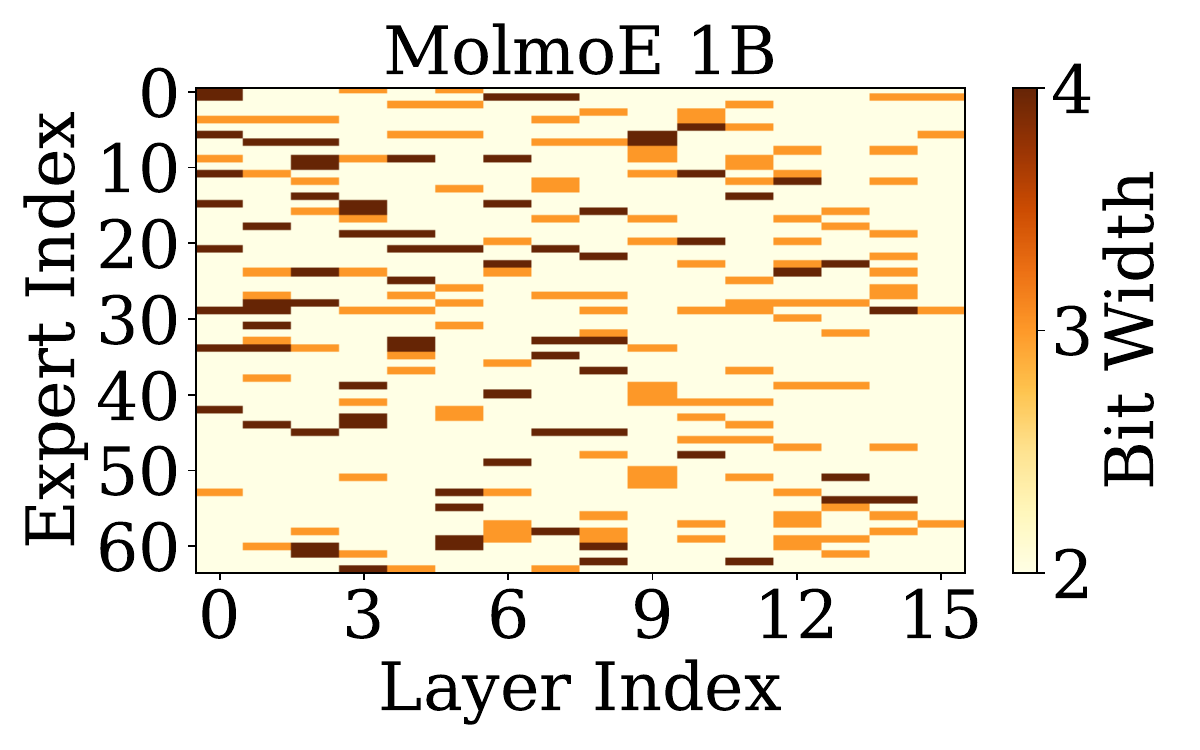}
            \label{subfig:molmoe_vl2_per_model_combined_3_clusters}
        }
        \subfloat[DeepSeek VL2-Tiny]{
            \includegraphics[width=.24\linewidth]{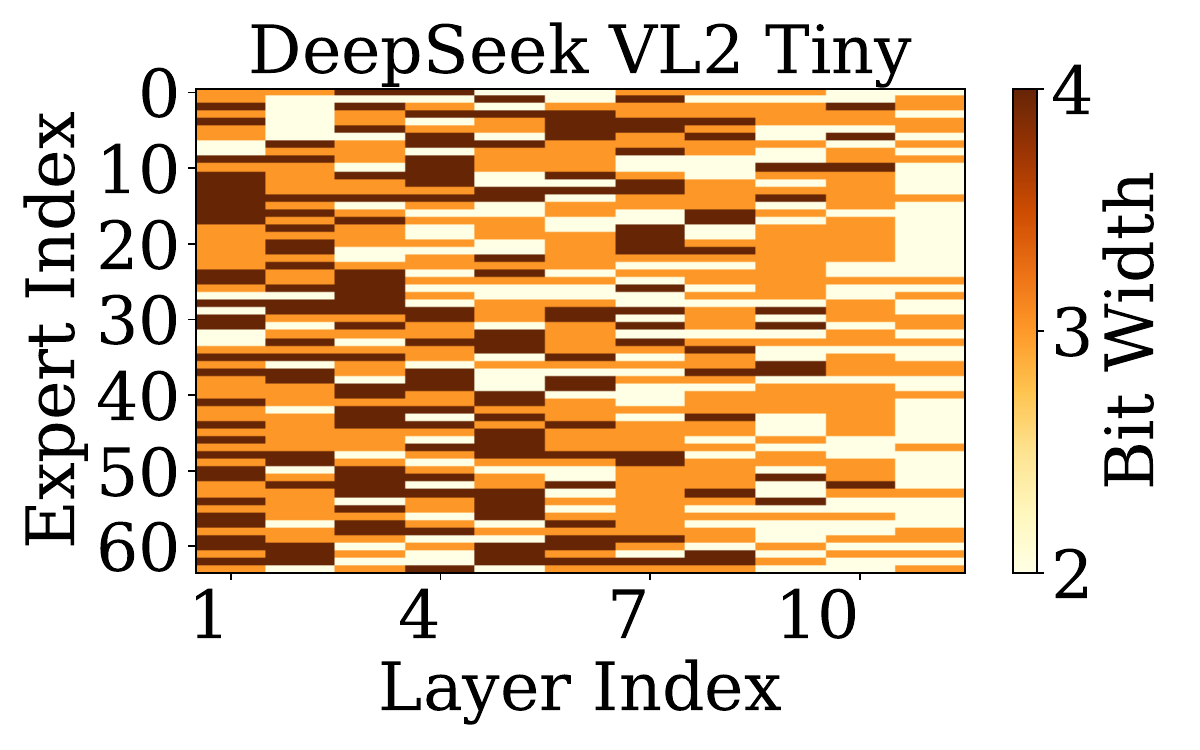}
            \label{subfig:ds_vl2_tiny_per_model_combined_3_clusters}
        }
        \subfloat[DeepSeek VL2-Small]{
            \includegraphics[width=.24\linewidth]{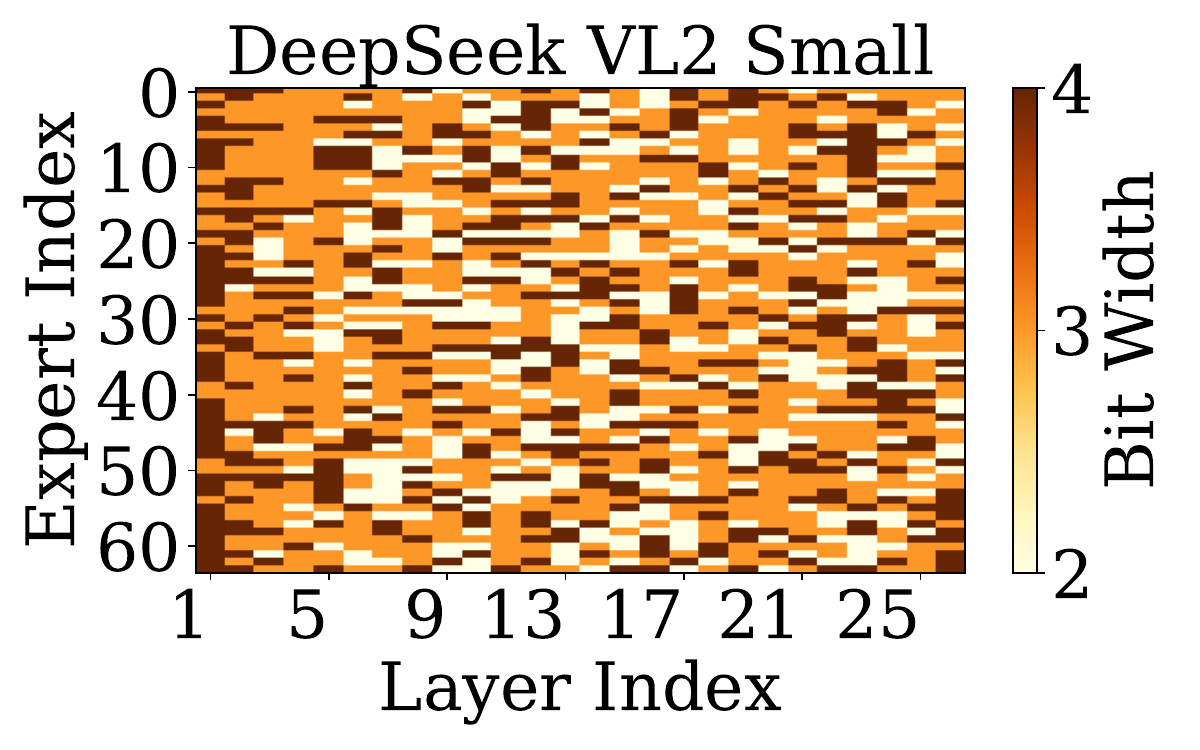}
            \label{subfig:ds_vl2_small_per_model_combined_3_clusters}
        }
        \subfloat[DeepSeek VL2-Base]{
            \includegraphics[width=.24\linewidth]{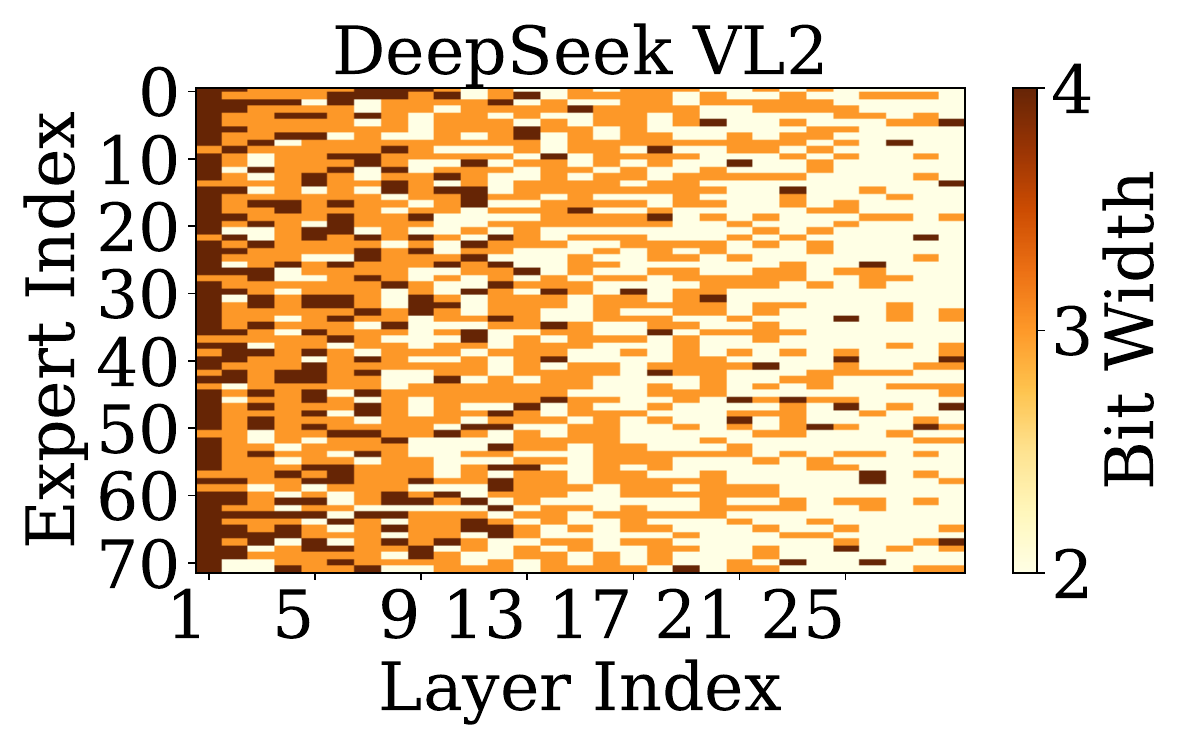}
            \label{subfig:ds_vl2_per_model_combined_3_clusters}
        }
        \vspace{-2mm}
        \caption{Model-wise Precision Assignment Map based on Normalized Hessian Trace \& Expert Activation Frequency}
        \vspace{-4mm}
        \label{fig:per_model_combined_3_clusters}
\end{figure*}

\section{MoPEQ}


\subsection{Expert Clustering}
Traditional mixed precision assignment methods \cite{dong2019hawq} typically allocate precisions to different layers based on predefined percentage splits without fully considering the relative importance of each layer. For example, if a model has 10 layers and 8 of them are highly critical but fall within a similar importance range, a rigid 50-50\% split would assign half of the layers to high precision and the other half to lower precision, regardless of their actual significance. This approach can lead to accuracy degradation, as some of the most important layers may be quantized to lower precision, negatively impacting the model quality.


\begin{algorithm}
\caption{MoPEQ: Precision Assignment via Expert Importance Clustering}
\label{Alg:mopeq}
\begin{algorithmic}[1]

\Require Experts $E = \{E_1, \dots, E_n\}$, Importance map $I: \mathcal{I} \to \mathbb{R}^+$, Bit widths $P = \{p_1, \dots, p_C\} = \{2,3,4\}$

\Require $C$ (Number of clusters) = len($P$)

\Ensure $A: P \to E$ (Allocation of Precision to Experts)

\State $V \gets \{ I(l) \mid l \in \mathcal{L} \}$ \Comment{Extract importance values}
\State Apply K-Means clustering on $V$ 


\For{each cluster $c \in \{1, \dots, C\}$}
    \State Compute cluster mean: $\mu_c = \frac{1}{|L_c|} \sum_{i \in L_c} V_i$
    
\EndFor


\State Sort clusters by $\mu_c$ in descending order $O = \{c_1, \dots, c_C\}$ and assign precisions: $A \gets \{O_i \to P_i\}_{1}^{C}$


\State \Return $A$
\end{algorithmic}
\end{algorithm}


Our MoPEQ approach (Algorithm \ref{Alg:mopeq}) utilizes K-means clustering to assign precisions to experts based on their relative importance. It first defines an assignment bit-width list, such as $\{p_1, p_2, p_3\} = \{2,3,4\}$. The goal is to partition the set of experts $\{E_1, E_2, \dots, E_n\}$ into $C$ clusters, where $C$ = len($P$). The clustering criterion is based on the importance metric,
ensuring that experts with similar importance values are grouped together by applying K-means clustering.
For each cluster $c$, we compute its mean importance as $\mu_c = \frac{1}{|L_c|} \sum_{i \in L_c} V_i$, where $V_i$ represents the importance value of expert $E_i$. The clusters are then sorted in descending order based on $\mu_c$.
Finally, we assign precision values such that the highest bit width is allocated to the cluster with the highest mean importance, and decreasing bit widths are assigned to the remaining clusters according to their rank, ensuring $A \leftarrow \{O_i \to P'_i\}_{i=1}^{C}$, where $P'$ is the precision list sorted in descending order. This approach effectively balances model efficiency and accuracy by ensuring that highly important experts receive higher bit widths while minimizing computational costs for less significant experts.

\subsection{Expert Bit width Assignment}

We investigate layer-wise and model-wise precision assignment using the expert clustering method. Layer-wise assignment, proposed in earlier studies \cite{huang2024mc}, groups experts within each layer based on their local importance. Figures \ref{fig:per_layer_freq_3_clusters} and \ref{fig:per_layer_hessian_3_clusters} illustrate the layer-wise bit width assignment based on expert activation frequency and hessian sensitivity, respectively. As depicted in these figures, this approach enables layer-specific optimization, allowing precision allocation to match layer expert activation patterns. However, this does not consider the impact of the layer on the overall model performance. Global model-wise clustering treats the entire model's experts as a single entity, grouping them based on the absolute importance of the layer. Figures \ref{fig:per_model_freq_3_clusters}, \ref{fig:per_layer_hessian_3_clusters} and \ref{fig:per_model_combined_3_clusters} illustrate the model-wise assignment map based on three importance metrics, activation frequency, hessian sensitivity and a normalized combination of frequency and sensitivity. This holistic prioritizes model-wide efficiency by allocating higher precision to layers with the most important experts, even if their layer-specific contributions appear modest. Specifically, we observe that initial layers receive higher precision while deeper layers receive lower bit width with Hessian importance. However, with activation frequency as the importance metric, we observe no significant difference in the precision allocated to experts between layer-wise and model-wise assignments. Activation frequency provides only localized information about how each expert is utilized relative to others within the same MoE layer. The sparse heatmaps across all models show that our approach effectively balances model size by preserving high precision in the most critical components.

\section{Evaluation}


\subsection{Experimental Setup}

We use the AutoRound framework \cite{intel_auto_round} to quantize VLM-MoEs to different precisions. We use SignRound \cite{cheng2023optimize} (summarized in Section \ref{sec:quant_methods}) as the quantization function and the baseline uniform quantization method. 
The primary contribution of our paper is expert precision assignment, enabling seamless integration into any quantization framework that supports the quantization of the given model.  
We compare our precision allocation method with activation frequency \cite{li2024examining} and layer-wise assignment scheme \cite{huang2024mc}. 
For VLM accuracy evaluation, we use the following datasets and tasks from VLMEvalKit \cite{duan2024vlmevalkit}: MME \cite{yin2023survey}, 
TextVQA \cite{singh2019towards}, AI2D \cite{kembhavi2016diagram}, DocVQA \cite{mathew2021docvqa}, MMMU \cite{yue2024mmmu}, InfoVQA \cite{mathew2022infographicvqa}, RealWorldQA \cite{zhang2024mme}, ScienceQA \cite{lu2022learn}. 
Our quantization search space encompasses bit widths of 2, 3, and 4 bits for mixed-precision assignment. We focus on these configurations because 4 bits are already considered state-of-the-art, and our objective is to explore beyond this threshold.
In Table \ref{tab:molmoe_results}, \ref{tab:ds_tiny_results}, \ref{tab:ds_small_results} and \ref{tab:ds_base_results}, we evaluate the validation performance and model size of our MoPEQ method against baseline uniform quantization and activation frequency-based mixed precision assignment across four VLM-MoEs.

\subsection{Activation Frequency vs Hessian Sensitivity}

DeepSeek-VL2-Base model, the largest model in our comparisons, achieves superior validation accuracy in seven out of nine tasks using our sensitivity approach compared to activation frequency. Using our sensitivity and model-wise precision assignment approach, the DeepSeek-VL2-Base model (10.485 GB) achieves superior accuracy on many tasks compared to uniform 4-bit and other baselines. This highlights the effectiveness of our methodology that as the model size grows, our method can reduce memory footprint without significant accuracy loss. From MolmoE-1B results, our sensitivity-based model-wise assignment outperforms the baseline in most tasks, while the hybrid approach maintains comparable accuracy with a smaller model size. The MolmoE-1B results clearly show our hybrid approach can be leveraged in load imbalance scenarios to reduce the model size by allocating higher precision only to experts which are both sensitive and frequently activated. Notably, in tasks like MME perception, our method surpasses the uniform 4-bit while requiring fewer parameters, demonstrating the strength of our sensitivity and hybrid strategies without relying on any dataset.

\subsection{Layer-wise vs Model-wise}

From the results in the Tables, we observe that model-wise precision allocation yields better validation performance across nine tasks on four models in 63 different scenarios compared to 42 scenarios of layer-wise allocation. This holds true even though the model sizes of mixed-precision quantized models remain similar under both model-wise and layer-wise schemes. Specifically, the DeepSeek-VL2 Small model, which differs slightly from other VLM-MoEs in our evaluation, presents an interesting case. Out of 27 possible combinations—comprising three importance metrics and nine tasks, model-wise assignment achieves better accuracy 22 times. The uniform distribution of expert frequency (Figure \ref{fig:activation_freq}(c)) and sensitivity (Figure \ref{fig:hessian_trace}(c)) across all experts and layers in this model complicates the quantization of experts to different precisions within a single layer. Consequently, when precision is unevenly distributed across layers, some layers can be heavily quantized while others remain largely unquantized.




\begin{table*}[h!]
\centering
\setlength{\tabcolsep}{2.9pt}
\renewcommand{\arraystretch}{1.1}
\caption{MolmoE-1B Model Performance on VLMEvalKit Tasks. 
The best validation accuracy for the mixed precision quantized models is in bold.
}
\resizebox{\textwidth}{!}{

\begin{tabular}{|l|c|c|c|c|c|c|c|c|c|c|c|c|}
\hline
\cellcolor[HTML]{C0C0C0}\begin{tabular}[c]{@{}l@{}}Quant.\\ Mthod\end{tabular}    & \cellcolor[HTML]{C0C0C0}\begin{tabular}[c]{@{}c@{}}Precision\\ Assignment\end{tabular} & \cellcolor[HTML]{C0C0C0}\begin{tabular}[c]{@{}c@{}}Expert\\ Importance\end{tabular} & \cellcolor[HTML]{C0C0C0}\begin{tabular}[c]{@{}c@{}}Expert\\ Precision\end{tabular} & \cellcolor[HTML]{FFFC9E}\begin{tabular}[c]{@{}c@{}}Model\\ Size (GB)\end{tabular} & \cellcolor[HTML]{CBCEFB}\begin{tabular}[c]{@{}c@{}}DocVQA\\ VAL\end{tabular}              & \cellcolor[HTML]{CBCEFB}\begin{tabular}[c]{@{}c@{}}InfoVQA\\ VAL\end{tabular}             & \cellcolor[HTML]{CBCEFB}\begin{tabular}[c]{@{}c@{}}MME-\\ Reasoning\end{tabular}            & \cellcolor[HTML]{CBCEFB}\begin{tabular}[c]{@{}c@{}}MME-\\ Perception\end{tabular}             & \cellcolor[HTML]{CBCEFB}\begin{tabular}[c]{@{}c@{}}MMMU\\ VAL\end{tabular}                & \cellcolor[HTML]{CBCEFB}RealWorldQA                                                      & \cellcolor[HTML]{CBCEFB}\begin{tabular}[c]{@{}c@{}}ScienceQA\\ VAL\end{tabular}           & \cellcolor[HTML]{CBCEFB}BLINK                                                             \\ \hline
                                                                                  & Uniform                                                                                &                                                                                     & 16                                                                                 & 13.45                                                                             & 76.605                                                            & 53.974                                                            & 265.71                                                              & 1360.67                                                               & 35.333                                                            & 58.562                                                           & 33.953                                                            & 40.978                                                            \\ \cline{2-2} \cline{4-13} 
                                                                                  & \begin{tabular}[c]{@{}c@{}}Uniform-\\ AutoRound\end{tabular}                           & \multirow{-2}{*}{Equal}                                                             & \begin{tabular}[c]{@{}c@{}}8\\ 4\end{tabular}                                      & \begin{tabular}[c]{@{}c@{}}7.20\\ 4.08\end{tabular}                               & \begin{tabular}[c]{@{}c@{}}76.574\\ 75.937\end{tabular}           & \begin{tabular}[c]{@{}c@{}}53.464\\ 53.587\end{tabular}           & \begin{tabular}[c]{@{}c@{}}260.0\\ 272.5\end{tabular}               & \begin{tabular}[c]{@{}c@{}}1368.128\\ 1300.088\end{tabular}           & \begin{tabular}[c]{@{}c@{}}35.333\\ 32.444\end{tabular}           & \begin{tabular}[c]{@{}c@{}}58.562\\ 57.386\end{tabular}          & \begin{tabular}[c]{@{}c@{}}34.049\\ 33.0\end{tabular}             & \begin{tabular}[c]{@{}c@{}}40.347\\ 40.558\end{tabular}           \\ \cline{2-13} 
\multirow{-3}{*}{Baseline}                                                        & \begin{tabular}[c]{@{}c@{}}Activation\\ Frequency\end{tabular}                         & \begin{tabular}[c]{@{}c@{}}Layer-wise\\ Model-wise\end{tabular}                     &                                                                                    & \begin{tabular}[c]{@{}c@{}}2.82\\ 2.80\end{tabular}                               & \begin{tabular}[c]{@{}c@{}}73.313\\ 73.137\end{tabular}           & \begin{tabular}[c]{@{}c@{}}48.246\\ 48.927\end{tabular}           & \begin{tabular}[c]{@{}c@{}}271.071\\ 234.285\end{tabular}           & \begin{tabular}[c]{@{}c@{}}1112.078\\ 1013.495\end{tabular}           & \begin{tabular}[c]{@{}c@{}}32.667\\ 31.778\end{tabular}           & \begin{tabular}[c]{@{}c@{}}54.51\\ 54.771\end{tabular}           & \begin{tabular}[c]{@{}c@{}}31.712\\ 31.283\end{tabular}           & \begin{tabular}[c]{@{}c@{}}38.874\\ 40.4\end{tabular}             \\ \cline{1-3} \cline{5-13} 
                                                                                  & \begin{tabular}[c]{@{}c@{}}Hessian\\ Sensitivity\end{tabular}                          & \begin{tabular}[c]{@{}c@{}}Layer-wise\\ Model-wise\end{tabular}                     &                                                                                    & \begin{tabular}[c]{@{}c@{}}3.31\\ 3.41\end{tabular}                               & \begin{tabular}[c]{@{}c@{}}75.206\\ \textbf{75.495}\end{tabular}           & \begin{tabular}[c]{@{}c@{}}49.897\\ \textbf{51.288}\end{tabular} & \begin{tabular}[c]{@{}c@{}}266.428\\ \textbf{291.071}\end{tabular} & \begin{tabular}[c]{@{}c@{}}1304.265\\ \textbf{1338.090}\end{tabular} & \begin{tabular}[c]{@{}c@{}}30.667\\ 33.333\end{tabular}           & \begin{tabular}[c]{@{}c@{}}55.163\\ \textbf{56.34}\end{tabular} & \begin{tabular}[c]{@{}c@{}}\textbf{32.427}\\ 32.141\end{tabular} & \begin{tabular}[c]{@{}c@{}}38.033\\ \textbf{41.399}\end{tabular} \\ \cline{2-3} \cline{5-13} 
\multirow{-2}{*}{\textbf{\begin{tabular}[c]{@{}l@{}}MoPEQ\\ (Ours)\end{tabular}}} & \begin{tabular}[c]{@{}c@{}}Normalized\\ Frequency-\\ Sensitivity\end{tabular}          & \begin{tabular}[c]{@{}c@{}}Layer-wise\\ Model-wise\end{tabular}                     & \multirow{-3}{*}{\begin{tabular}[c]{@{}c@{}}Mixed-\\ 2,3,4\end{tabular}}           & \begin{tabular}[c]{@{}c@{}}\textbf{2.77}\\ 2.79\end{tabular}                     & \begin{tabular}[c]{@{}c@{}}72.526\\ 72.668\end{tabular} & \begin{tabular}[c]{@{}c@{}}49.007\\ 48.41\end{tabular}            & \begin{tabular}[c]{@{}c@{}}249.286\\ 257.143\end{tabular}           & \begin{tabular}[c]{@{}c@{}}1019.842\\ 1245.821\end{tabular}           & \begin{tabular}[c]{@{}c@{}}\textbf{35.333}\\ 30.778\end{tabular} & \begin{tabular}[c]{@{}c@{}}53.856\\ 52.026\end{tabular}          & \begin{tabular}[c]{@{}c@{}}32.332\\ 30.567\end{tabular}           & \begin{tabular}[c]{@{}c@{}}39.663\\ 38.927\end{tabular}           \\ \hline
\end{tabular}

}
        
        \label{tab:molmoe_results}
    \end{table*}

\begin{table*}[h!]
\centering
\setlength{\tabcolsep}{2.9pt}
\renewcommand{\arraystretch}{1.1}
\caption{DeepSeek-VL2-Tiny Model Performance on VLMEvalKit Tasks. 
}
\resizebox{\textwidth}{!}{

\begin{tabular}{|l|c|c|c|c|c|c|c|c|c|c|c|c|c|}
\hline
\cellcolor[HTML]{C0C0C0}\begin{tabular}[c]{@{}l@{}}Quant.\\ Method\end{tabular}   & \cellcolor[HTML]{C0C0C0}\begin{tabular}[c]{@{}c@{}}Precision\\ Assignment\end{tabular} & \cellcolor[HTML]{C0C0C0}\begin{tabular}[c]{@{}c@{}}Expert\\ Importance\end{tabular} & \cellcolor[HTML]{C0C0C0}\begin{tabular}[c]{@{}c@{}}Expert\\ Precision\end{tabular} & \cellcolor[HTML]{FFFC9E}\begin{tabular}[c]{@{}c@{}}Model\\ Size (GB)\end{tabular} & \cellcolor[HTML]{CBCEFB}\begin{tabular}[c]{@{}c@{}}AI2D TEST\\ NO MASK\end{tabular}       & \cellcolor[HTML]{CBCEFB}\begin{tabular}[c]{@{}c@{}}DocVQA\\ VAL\end{tabular}             & \cellcolor[HTML]{CBCEFB}\begin{tabular}[c]{@{}c@{}}InfoVQA\\ VAL\end{tabular}             & \cellcolor[HTML]{CBCEFB}\begin{tabular}[c]{@{}c@{}}MME-\\ Reasoning\end{tabular}          & \cellcolor[HTML]{CBCEFB}\begin{tabular}[c]{@{}c@{}}MME-\\ Perception\end{tabular}             & \cellcolor[HTML]{CBCEFB}\begin{tabular}[c]{@{}c@{}}MMMU\\ VAL\end{tabular}              & \cellcolor[HTML]{CBCEFB}RealWorldQA                                                       & \cellcolor[HTML]{CBCEFB}\begin{tabular}[c]{@{}c@{}}ScienceQA\\ VAL\end{tabular}           & \cellcolor[HTML]{CBCEFB}BLINK                                                             \\ \hline
                                                                                  & Uniform                                                                                &                                                                                     & 16                                                                                 & 6.278                                                                             & 84.326                                                            & 88.467                                                           & 63.892                                                            & 364.643                                                           & 1554.758                                                              & 40.222                                                          & 65.49                                                             & 88.46                                                             & 40.505                                                            \\ \cline{2-2} \cline{4-14} 
                                                                                  & \begin{tabular}[c]{@{}c@{}}Uniform-\\ AutoRound\end{tabular}                           & \multirow{-2}{*}{Equal}                                                             & \begin{tabular}[c]{@{}c@{}}8\\ 4\end{tabular}                                      & \begin{tabular}[c]{@{}c@{}}3.854\\ 2.642\end{tabular}                             & \begin{tabular}[c]{@{}c@{}}84.294\\ 83.549\end{tabular}           & \begin{tabular}[c]{@{}c@{}}88.632\\ 88.162\end{tabular}          & \begin{tabular}[c]{@{}c@{}}64.003\\ 62.959\end{tabular}           & \begin{tabular}[c]{@{}c@{}}357.143\\ 357.857\end{tabular}         & \begin{tabular}[c]{@{}c@{}}1549.785\\ 1563.248\end{tabular}           & \begin{tabular}[c]{@{}c@{}}40.111\\ 39.111\end{tabular}         & \begin{tabular}[c]{@{}c@{}}65.229\\ 65.098\end{tabular}           & \begin{tabular}[c]{@{}c@{}}88.317\\ 87.458\end{tabular}           & \begin{tabular}[c]{@{}c@{}}40.768\\ 42.083\end{tabular}           \\ \cline{2-14} 
\multirow{-3}{*}{Baseline}                                                        & \begin{tabular}[c]{@{}c@{}}Activation\\ Frequency\end{tabular}                         & \begin{tabular}[c]{@{}c@{}}Layer-wise\\ Model-wise\end{tabular}                     &                                                                                    & \begin{tabular}[c]{@{}c@{}}2.427\\ 2.410\end{tabular}                             & \begin{tabular}[c]{@{}c@{}}\textbf{84.003}\\ 83.776\end{tabular} & \begin{tabular}[c]{@{}c@{}}87.652\\ 87.666\end{tabular}          & \begin{tabular}[c]{@{}c@{}}61.955\\ \textbf{62.004}\end{tabular} & \begin{tabular}[c]{@{}c@{}}358.214\\ 340.0\end{tabular}           & \begin{tabular}[c]{@{}c@{}}1561.679\\ 1561.810\end{tabular}           & \begin{tabular}[c]{@{}c@{}}37.111\\ 38.111\end{tabular}         & \begin{tabular}[c]{@{}c@{}}\textbf{66.144}\\ 66.013\end{tabular} & \begin{tabular}[c]{@{}c@{}}86.934\\ 86.552\end{tabular}           & \begin{tabular}[c]{@{}c@{}}41.189\\ 42.188\end{tabular}           \\ \cline{1-3} \cline{5-14} 
                                                                                  & \begin{tabular}[c]{@{}c@{}}Hessian\\ Sensitivity\end{tabular}                          & \begin{tabular}[c]{@{}c@{}}Layer-wise\\ Model-wise\end{tabular}                     &                                                                                    & \begin{tabular}[c]{@{}c@{}}2.367\\ 2.349\end{tabular}                             & \begin{tabular}[c]{@{}c@{}}83.97\\ 83.323\end{tabular}            & \begin{tabular}[c]{@{}c@{}}87.334\\ 87.633\end{tabular}          & \begin{tabular}[c]{@{}c@{}}61.645\\ 60.983\end{tabular}           & \begin{tabular}[c]{@{}c@{}}371.786\\ 374.643\end{tabular}         & \begin{tabular}[c]{@{}c@{}}1561.112\\ 1560.403\end{tabular}           & \begin{tabular}[c]{@{}c@{}}\textbf{38.556}\\ 32.0\end{tabular} & \begin{tabular}[c]{@{}c@{}}63.791\\ 64.837\end{tabular}           & \begin{tabular}[c]{@{}c@{}}87.172\\ \textbf{87.458}\end{tabular} & \begin{tabular}[c]{@{}c@{}}\textbf{42.609}\\ 40.505\end{tabular} \\ \cline{2-3} \cline{5-14} 
\multirow{-2}{*}{\textbf{\begin{tabular}[c]{@{}l@{}}MoPEQ\\ (Ours)\end{tabular}}} & \begin{tabular}[c]{@{}c@{}}Normalized\\ Frequency-\\ Sensitivity\end{tabular}          & \begin{tabular}[c]{@{}c@{}}Layer-wise\\ Model-wise\end{tabular}                     & \multirow{-3}{*}{\begin{tabular}[c]{@{}c@{}}Mixed-\\ 2,3,4\end{tabular}}           & \begin{tabular}[c]{@{}c@{}}\textbf{2.311}\\ 2.348\end{tabular}                   & \begin{tabular}[c]{@{}c@{}}83.258\\ 83.355\end{tabular}           & \begin{tabular}[c]{@{}c@{}}87.106\\ \textbf{87.95}\end{tabular} & \begin{tabular}[c]{@{}c@{}}60.858\\ 61.695\end{tabular}           & \begin{tabular}[c]{@{}c@{}}370.0\\ \textbf{376.786}\end{tabular} & \begin{tabular}[c]{@{}c@{}}1532.439\\ \textbf{1585.692}\end{tabular} & \begin{tabular}[c]{@{}c@{}}38.111\\ 33.333\end{tabular}         & \begin{tabular}[c]{@{}c@{}}64.052\\ 64.575\end{tabular}           & \begin{tabular}[c]{@{}c@{}}86.028\\ 87.029\end{tabular}           & \begin{tabular}[c]{@{}c@{}}42.188\\ 40.768\end{tabular}           \\ \hline
\end{tabular}

}
        
        \label{tab:ds_tiny_results}
    \end{table*}

\begin{table*}[h!]
\centering
\setlength{\tabcolsep}{2.9pt}
\renewcommand{\arraystretch}{1.1}
\caption{DeepSeek-VL2-Small Model Performance on VLMEvalKit Tasks. 
}
\resizebox{\textwidth}{!}{

\begin{tabular}{|l|c|c|c|c|c|c|c|c|c|c|c|c|c|}
\hline
\cellcolor[HTML]{C0C0C0}\begin{tabular}[c]{@{}l@{}}Quant.\\ Method\end{tabular}   & \cellcolor[HTML]{C0C0C0}\begin{tabular}[c]{@{}c@{}}Precision\\ Assignment\end{tabular} & \cellcolor[HTML]{C0C0C0}\begin{tabular}[c]{@{}c@{}}Expert\\ Importance\end{tabular} & \cellcolor[HTML]{C0C0C0}\begin{tabular}[c]{@{}c@{}}Expert\\ Precision\end{tabular} & \cellcolor[HTML]{FFFC9E}\begin{tabular}[c]{@{}c@{}}Model\\ Size (GB)\end{tabular} & \cellcolor[HTML]{CBCEFB}\begin{tabular}[c]{@{}c@{}}AI2D TEST\\ NO MASK\end{tabular}                       & \cellcolor[HTML]{CBCEFB}\begin{tabular}[c]{@{}c@{}}DocVQA\\ VAL\end{tabular}                            & \cellcolor[HTML]{CBCEFB}\begin{tabular}[c]{@{}c@{}}InfoVQA\\ VAL\end{tabular}                           & \cellcolor[HTML]{CBCEFB}\begin{tabular}[c]{@{}c@{}}MME-\\ Reasoning\end{tabular}                          & \cellcolor[HTML]{CBCEFB}\begin{tabular}[c]{@{}c@{}}MME-\\ Perception\end{tabular}                             & \cellcolor[HTML]{CBCEFB}\begin{tabular}[c]{@{}c@{}}MMMU\\ VAL\end{tabular}                               & \cellcolor[HTML]{CBCEFB}RealWorldQA                                                                       & \cellcolor[HTML]{CBCEFB}\begin{tabular}[c]{@{}c@{}}ScienceQA\\ VAL\end{tabular}                           & \cellcolor[HTML]{CBCEFB}BLINK                                                                             \\ \hline
                                                                                  & Uniform                                                                                &                                                                                     & 16                                                                                 & 30.078                                                                            & 88.342                                                                            & 92.296                                                                          & 72.38                                                                           & 513.214                                                                           & 1649.135                                                                              & 48.0                                                                             & 69.804                                                                            & 94.373                                                                            & 52.236                                                                            \\ \cline{2-2} \cline{4-14} 
                                                                                  & \begin{tabular}[c]{@{}c@{}}Uniform-\\ AutoRound\end{tabular}                           & \multirow{-2}{*}{Equal}                                                             & \begin{tabular}[c]{@{}c@{}}8\\ 4\end{tabular}                                      & \begin{tabular}[c]{@{}c@{}}15.845\\ 8.728\end{tabular}                            & \begin{tabular}[c]{@{}c@{}}73.154\\ 71.503\end{tabular}                           & \begin{tabular}[c]{@{}c@{}}63.372\\ 62.58\end{tabular}                             & \begin{tabular}[c]{@{}c@{}}44.328\\ 42.274\end{tabular}                         & \begin{tabular}[c]{@{}c@{}}494.643\\ 485.357\end{tabular}                         & \begin{tabular}[c]{@{}c@{}}1643.240\\ 1640.147\end{tabular}                           & \begin{tabular}[c]{@{}c@{}}33.222\\ 34.0\end{tabular}                            & \begin{tabular}[c]{@{}c@{}}58.431\\ 57.124\end{tabular}                           & \begin{tabular}[c]{@{}c@{}}94.564\\ 95.279\end{tabular}                           & \begin{tabular}[c]{@{}c@{}}42.557\\ 41.82\end{tabular}                            \\ \cline{2-14} 
\multirow{-3}{*}{Baseline}                                                        & \begin{tabular}[c]{@{}c@{}}Activation\\ Frequency\end{tabular}                         & \begin{tabular}[c]{@{}c@{}}Layer-wise\\ Model-wise\end{tabular}                     &                                                                                    & \begin{tabular}[c]{@{}c@{}}7.357\\ 7.305\end{tabular}                             & \begin{tabular}[c]{@{}c@{}}51.036\\ \textbf{70.531}\end{tabular} & \begin{tabular}[c]{@{}c@{}}13.193\\ 62.005\end{tabular}                         & \begin{tabular}[c]{@{}c@{}}24.568\\ 41.174\end{tabular}                         & \begin{tabular}[c]{@{}c@{}}372.857\\ \textbf{486.071}\end{tabular} & \begin{tabular}[c]{@{}c@{}}1532.495\\ \textbf{1678.309}\end{tabular} & \begin{tabular}[c]{@{}c@{}}9.333\\ \textbf{32.556}\end{tabular} & \begin{tabular}[c]{@{}c@{}}43.007\\ 55.425\end{tabular}                           & \begin{tabular}[c]{@{}c@{}}85.312\\ \textbf{95.136}\end{tabular} & \begin{tabular}[c]{@{}c@{}}38.611\\ 41.189\end{tabular}                           \\ \cline{1-3} \cline{5-14} 
                                                                                  & \begin{tabular}[c]{@{}c@{}}Hessian\\ Sensitivity\end{tabular}                          & \begin{tabular}[c]{@{}c@{}}Layer-wise\\ Model-wise\end{tabular}                     &                                                                                    & \begin{tabular}[c]{@{}c@{}}7.215\\ 7.141\end{tabular}                             & \begin{tabular}[c]{@{}c@{}}41.767\\ 69.56\end{tabular}                            & \begin{tabular}[c]{@{}c@{}}62.018\\ 61.737\end{tabular}                         & \begin{tabular}[c]{@{}c@{}}40.589\\ 40.054\end{tabular}                         & \begin{tabular}[c]{@{}c@{}}481.071\\ 466.786\end{tabular}                         & \begin{tabular}[c]{@{}c@{}}1675.200\\ 1659.337\end{tabular}                           & \begin{tabular}[c]{@{}c@{}}30.0\\ 32.0\end{tabular}                              & \begin{tabular}[c]{@{}c@{}}\textbf{56.601}\\ 55.425\end{tabular} & \begin{tabular}[c]{@{}c@{}}93.515\\ 93.562\end{tabular}                           & \begin{tabular}[c]{@{}c@{}}41.767\\ \textbf{42.767}\end{tabular} \\ \cline{2-3} \cline{5-14} 
\multirow{-2}{*}{\textbf{\begin{tabular}[c]{@{}l@{}}MoPEQ\\ (Ours)\end{tabular}}} & \begin{tabular}[c]{@{}c@{}}Normalized\\ Frequency-\\ Sensitivity\end{tabular}          & \begin{tabular}[c]{@{}c@{}}Layer-wise\\ Model-wise\end{tabular}                     & \multirow{-3}{*}{\begin{tabular}[c]{@{}c@{}}Mixed-\\ 2,3,4\end{tabular}}           & \begin{tabular}[c]{@{}c@{}}\textbf{6.941}\\ 7.113\end{tabular}   & \begin{tabular}[c]{@{}c@{}}55.019\\ 70.045\end{tabular}                           & \begin{tabular}[c]{@{}c@{}}14.194\\ \textbf{62.356}\end{tabular} & \begin{tabular}[c]{@{}c@{}}24.255\\ \textbf{41.169}\end{tabular} & \begin{tabular}[c]{@{}c@{}}372.857\\ 485.714\end{tabular}                         & \begin{tabular}[c]{@{}c@{}}1552.618\\ 1666.1694\end{tabular}                          & \begin{tabular}[c]{@{}c@{}}18.889\\ 30.667\end{tabular}                          & \begin{tabular}[c]{@{}c@{}}46.797\\ 55.163\end{tabular}                           & \begin{tabular}[c]{@{}c@{}}87.935\\ 94.278\end{tabular}                           & \begin{tabular}[c]{@{}c@{}}39.663\\ 42.083\end{tabular}                           \\ \hline
\end{tabular}

}
        
        \label{tab:ds_small_results}
    \end{table*}
\begin{table*}[h!]
\centering
\setlength{\tabcolsep}{2.9pt}
\renewcommand{\arraystretch}{1.1}
\caption{DeepSeek-VL2 Model Performance on VLMEvalKit Tasks. 
}
\resizebox{\textwidth}{!}{

\begin{tabular}{|l|c|c|c|c|c|c|c|c|c|c|c|c|c|}
\hline
\cellcolor[HTML]{C0C0C0}\begin{tabular}[c]{@{}l@{}}Quant.\\ Method\end{tabular}   & \cellcolor[HTML]{C0C0C0}\begin{tabular}[c]{@{}c@{}}Precision\\ Assignment\end{tabular} & \cellcolor[HTML]{C0C0C0}\begin{tabular}[c]{@{}c@{}}Expert\\ Importance\end{tabular} & \cellcolor[HTML]{C0C0C0}\begin{tabular}[c]{@{}c@{}}Expert\\ Precision\end{tabular} & \cellcolor[HTML]{FFFC9E}\begin{tabular}[c]{@{}c@{}}Model\\ Size (GB)\end{tabular} & \cellcolor[HTML]{CBCEFB}\begin{tabular}[c]{@{}c@{}}AI2D TEST\\ NO MASK\end{tabular}       & \cellcolor[HTML]{CBCEFB}\begin{tabular}[c]{@{}c@{}}DocVQA\\ VAL\end{tabular}              & \cellcolor[HTML]{CBCEFB}\begin{tabular}[c]{@{}c@{}}InfoVQA\\ VAL\end{tabular}             & \cellcolor[HTML]{CBCEFB}\begin{tabular}[c]{@{}c@{}}MME-\\ Reasoning\end{tabular}            & \cellcolor[HTML]{CBCEFB}\begin{tabular}[c]{@{}c@{}}MME-\\ Perception\end{tabular}             & \cellcolor[HTML]{CBCEFB}\begin{tabular}[c]{@{}c@{}}MMMU\\ VAL\end{tabular}              & \cellcolor[HTML]{CBCEFB}RealWorldQA                                                       & \cellcolor[HTML]{CBCEFB}\begin{tabular}[c]{@{}c@{}}ScienceQA\\ VAL\end{tabular}         & \cellcolor[HTML]{CBCEFB}BLINK                                                             \\ \hline
                                                                                  & Uniform                                                                                &                                                                                     & 16                                                                                 & 51.186                                                                            & 91.159                                                            & 93.184                                                            & 74.167                                                            & 594.643                                                             & 1641.053                                                              & 51.556                                                          & 70.588                                                            & 96.996                                                          & 54.866                                                            \\ \cline{2-2} \cline{4-14} 
                                                                                  & \begin{tabular}[c]{@{}c@{}}Uniform-\\ AutoRound\end{tabular}                           & \multirow{-2}{*}{Equal}                                                             & \begin{tabular}[c]{@{}c@{}}8\\ 4\end{tabular}                                      & \begin{tabular}[c]{@{}c@{}}26.630\\ 14.353\end{tabular}                           & \begin{tabular}[c]{@{}c@{}}76.49\\ 76.263\end{tabular}            & \begin{tabular}[c]{@{}c@{}}78.618\\ 77.397\end{tabular}           & \begin{tabular}[c]{@{}c@{}}57.652\\ 56.2\end{tabular}             & \begin{tabular}[c]{@{}c@{}}585.714\\ 539.286\end{tabular}           & \begin{tabular}[c]{@{}c@{}}1628.288\\ 1614.607\end{tabular}           & \begin{tabular}[c]{@{}c@{}}30.667\\ 37.667\end{tabular}         & \begin{tabular}[c]{@{}c@{}}62.222\\ 61.699\end{tabular}           & \begin{tabular}[c]{@{}c@{}}96.948\\ 96.614\end{tabular}         & \begin{tabular}[c]{@{}c@{}}47.449\\ 46.66\end{tabular}            \\ \cline{2-14} 
\multirow{-3}{*}{Baseline}                                                        & \begin{tabular}[c]{@{}c@{}}Activation\\ Frequency\end{tabular}                         & \begin{tabular}[c]{@{}c@{}}Layer-wise\\ Model-wise\end{tabular}                     &                                                                                    & \begin{tabular}[c]{@{}c@{}}11.374\\ 11.044\end{tabular}                           & \begin{tabular}[c]{@{}c@{}}75.324\\ \textbf{75.712}\end{tabular} & \begin{tabular}[c]{@{}c@{}}76.669\\ 76.055\end{tabular}           & \begin{tabular}[c]{@{}c@{}}55.284\\ 54.74\end{tabular}            & \begin{tabular}[c]{@{}c@{}}571.429\\ 564.286\end{tabular}           & \begin{tabular}[c]{@{}c@{}}1596.320\\ 1611.655\end{tabular}           & \begin{tabular}[c]{@{}c@{}}36.0\\ 36.667\end{tabular}           & \begin{tabular}[c]{@{}c@{}}61.699\\ 62.876\end{tabular}           & \begin{tabular}[c]{@{}c@{}}96.09\\ 96.233\end{tabular}          & \begin{tabular}[c]{@{}c@{}}46.239\\ 45.818\end{tabular}           \\ \cline{1-3} \cline{5-14} 
                                                                                  & \begin{tabular}[c]{@{}c@{}}Hessian\\ Sensitivity\end{tabular}                          & \begin{tabular}[c]{@{}c@{}}Layer-wise\\ Model-wise\end{tabular}                     &                                                                                    & \begin{tabular}[c]{@{}c@{}}11.554\\ \textbf{10.485}\end{tabular}                 & \begin{tabular}[c]{@{}c@{}}75.032\\ 75.389\end{tabular}           & \begin{tabular}[c]{@{}c@{}}\textbf{76.985}\\ 76.341\end{tabular} & \begin{tabular}[c]{@{}c@{}}55.054\\ \textbf{55.555}\end{tabular} & \begin{tabular}[c]{@{}c@{}}\textbf{578.571}\\ 576.426\end{tabular} & \begin{tabular}[c]{@{}c@{}}1623.521\\ 1623.449\end{tabular}           & \begin{tabular}[c]{@{}c@{}}37.0\\ 37.444\end{tabular}           & \begin{tabular}[c]{@{}c@{}}\textbf{63.268}\\ 61.176\end{tabular} & \begin{tabular}[c]{@{}c@{}}96.09\\ \textbf{96.28}\end{tabular} & \begin{tabular}[c]{@{}c@{}}\textbf{46.923}\\ 45.502\end{tabular} \\ \cline{2-3} \cline{5-14} 
\multirow{-2}{*}{\textbf{\begin{tabular}[c]{@{}l@{}}MoPEQ\\ (Ours)\end{tabular}}} & \begin{tabular}[c]{@{}c@{}}Normalized\\ Frequency-\\ Sensitivity\end{tabular}          & \begin{tabular}[c]{@{}c@{}}Layer-wise\\ Model-wise\end{tabular}                     & \multirow{-3}{*}{\begin{tabular}[c]{@{}c@{}}Mixed-\\ 2,3,4\end{tabular}}           & \begin{tabular}[c]{@{}c@{}}10.582\\ 10.664\end{tabular}                           & \begin{tabular}[c]{@{}c@{}}75.421\\ 74.903\end{tabular}           & \begin{tabular}[c]{@{}c@{}}76.308\\ 75.945\end{tabular}           & \begin{tabular}[c]{@{}c@{}}54.313\\ 54.649\end{tabular}           & \begin{tabular}[c]{@{}c@{}}573.929\\ 547.5\end{tabular}             & \begin{tabular}[c]{@{}c@{}}1607.808\\ \textbf{1627.548}\end{tabular} & \begin{tabular}[c]{@{}c@{}}\textbf{37.889}\\ 32.0\end{tabular} & \begin{tabular}[c]{@{}c@{}}61.961\\ 61.699\end{tabular}           & \begin{tabular}[c]{@{}c@{}}95.947\\ 95.851\end{tabular}         & \begin{tabular}[c]{@{}c@{}}45.818\\ 45.239\end{tabular}           \\ \hline
\end{tabular}

}
        
        \label{tab:ds_base_results}
    \end{table*}
    \vspace{2mm}

\subsection{Implications on Hardware Performance}
Although we did not include hardware performance results in our work due to the limitation of mixed precision quantization support for VLM-MoEs in existing inference frameworks (e.g., vLLM \cite{kwon2023efficient}), we briefly highlight the advantages of our MoPEQ approach. 
Activation frequency-based quantization assigns higher precision to frequently used experts, increasing computation time and communication overhead in memory-constrained environments with model offloading strategy. In contrast, the MoPEQ approach assigns lower precision even to frequently activated experts, reducing the data transferred between CPU and GPU, lowering communication overhead, and decreasing computation time. This results in improved overall inference efficiency due to reduced communication and computational costs.
In future work, we plan to integrate mixed-precision quantization support into vLLM and provide hardware performance evaluations.


\section{Conclusion}

In this work, we introduce MoPEQ, a novel post-training mixed precision quantization algorithm for VLM-MoEs that allocates bit widths based on expert sensitivity using Hessian trace approximation. Unlike existing methods that rely on activation frequency, our approach effectively balances accuracy and model size without requiring a calibration dataset to determine the importance of each expert. Through extensive experiments on state-of-the-art VLMs MolmoE-1B and DeepSeek VL2 variants, we demonstrated that MoPEQ achieves substantial memory savings while maintaining competitive accuracy across multiple VLMEvalKit tasks. For example, on the standard MME perception task, our sensitivity-based method achieves a score of 1338, surpassing the uniform 4-bit quantized model’s score of 1300 with a smaller model size. Our findings highlight the importance of expert sensitivity in MoE quantization over the repetitive use of an expert.

\section*{Acknowledgements}

This research used resources of the Argonne Leadership
Computing Facility, a U.S. Department of Energy (DOE)
Office of Science user facility at Argonne National Laboratory
and is based on research supported by the U.S. DOE Office
of Science-Advanced Scientific Computing Research Program,
under Contract No. DE-AC02-06CH11357

{
    \small
    \bibliographystyle{ieeenat_fullname}
    \bibliography{main}
}

\end{document}